\newcommand{\schemename}{SmartGen}
\documentclass[sigconf]{acmart}
\AtBeginDocument{%
  \providecommand\BibTeX{{%
    \normalfont B\kern-0.5em{\scshape i\kern-0.25em b}\kern-0.8em\TeX}}}

\setcopyright{acmlicensed}
\copyrightyear{2025}
\acmYear{2025}
\acmDOI{XXXXXXX.XXXXXXX}

\usepackage{algorithmic}
\usepackage[ruled, linesnumbered]{algorithm2e}
\usepackage{subfigure}
\usepackage{multirow}

\acmISBN{978-1-4503-XXXX-X/18/06}

\begin{document}

\newcommand{\timesplit}{Time and Semantic-aware Split}
\newcommand{\TS}{TSS}
\newcommand{\compression}{Semantic-aware Sequence Compression}
\newcommand{\cp}{SSC}
\newcommand{\graph}{Graph-guided Sequence Synthesis}
\newcommand{\gp}{GSS}
\newcommand{\filter}{Two-stage Outlier Filter}
\newcommand{\dof}{TOF}

% \setlength{\parskip}{0.1em}

% \title{Make Your Home Safe: Context-aware Unsupervised User Behavior Anomaly Detection in Smart Homes}

% \title{\schemename: Semantic-aware Graph-guided Behavior Sequences Generation with Large Language Models for Smart Homes}

\title{Semantic-aware Graph-guided Behavior Sequences Generation with Large Language Models for Smart Homes}
% \title{Make Model Work Well: A Efficient Model Adaptation System Driven by Large Language Model Synthetic Data for Smart Homes}
% \title{SmartGen: Synthesizing Context-Aware User Behavior Data for Adaptive Smart Home Intelligence}


% \author{Anonymous author(s)}

\author{Zhiyao Xu}
\affiliation{
  \institution{Tsinghua Shenzhen International Graduate School}
  \city{Shenzhen}
  \country{China}}
\email{zhixu9557@gmail.com}

\author{Dan Zhao}
\affiliation{
  \institution{Peng Cheng Laboratory}
  \city{Shenzhen}
  \country{China}}
\email{zhaod01@pcl.ac.cn}
\authornotemark[1]

\author{Qingsong Zou}
\affiliation{
  \institution{Tsinghua Shenzhen International Graduate School}
  \institution{Peng Cheng Laboratory}
  \city{Shenzhen}
  \country{China}}
\email{zouqs21@mails.tsinghua.edu.cn
}

\author{Qing Li}
\affiliation{
  \institution{Peng Cheng Laboratory}
  \city{Shenzhen}
  \country{China}}
\email{liq@pcl.ac.cn}

\author{Yong Jiang}
\affiliation{
  \institution{Tsinghua Shenzhen International Graduate School}
  \institution{Peng Cheng Laboratory}
  \city{Shenzhen}
  \country{China}}
\email{jiangy@sz.tsinghua.edu.cn}
\authornote{Corresponding authors.}

\author{Yuhang Wang}
\affiliation{
  \institution{Southwest University}
  \city{Chongqing}
  \country{China}}
\email{wyh20030323@email.swu.edu.cn}

\author{Jingyu Xiao}
\affiliation{
  \institution{The Chinese University of Hong Kong}
  \city{Hong Kong}
  \country{China}}
\email{jyxiao@link.cuhk.edu.hk}

% \affiliation{%

% \affiliation{%



\begin{abstract}

As smart homes become increasingly prevalent, intelligent models are widely used for tasks such as anomaly detection and behavior prediction. These models are typically trained on static datasets, making them brittle to behavioral drift caused by seasonal changes, lifestyle shifts, or evolving routines. However, collecting new behavior data for retraining is often impractical due to its slow pace, high cost, and privacy concerns. In this paper, we propose SmartGen, an LLM-based framework that synthesizes context-aware user behavior data to support continual adaptation of downstream smart home models. SmartGen consists of four key components. First, we design a Time and Semantic-aware Split module to divide long behavior sequences into manageable, semantically coherent subsequences under dual time-span constraints. Second, we propose Semantic-aware Sequence Compression to reduce input length while preserving representative semantics by clustering behavior mapping in latent space. Third, we introduce Graph-guided Sequence Synthesis, which constructs a behavior relationship graph and encodes frequent transitions into prompts, guiding the LLM to generate data aligned with contextual changes while retaining core behavior patterns. Finally, we design a Two-stage Outlier Filter to identify and remove implausible or semantically inconsistent outputs, aiming to improve the factual coherence and behavioral validity of the generated sequences. Experiments on three real-world datasets demonstrate that SmartGen significantly enhances model performance on anomaly detection and behavior prediction tasks under behavioral drift, with anomaly detection improving by 85.43\% and behavior prediction by 70.51\% on average. The code is available at \url{https://github.com/horizonsinzqs/SmartGen}.

\end{abstract}

% Due to the 
% some methods are proposd to xxx. However, temporal context, unbalance user behavior and noise behavior bring huge challenges for user modeling in smart homes. 





% \keywords{User Behavior Modeling, Anomaly Detection, Self-supervised Learning.}



\maketitle

\section{Introduction}
\label{sec:intro}

% With the development of 

% However, there are

%The rapid growth of Internet-of-Things (IoT) solutions has driven an unprecedented rise in the number of smart devices within homes, with estimates suggesting that this number will reach approximately 5 billion by 2025~\cite{iot-analytics}. 
The IoT boom has spurred exponential growth in smart home devices, projected to hit 5 billion by 2025~\cite{iot-analytics}. In smart homes, IoT devices monitor users' living environments, execut users' instructions and interact directly with their living spaces. 
To further realize the whole house intelligence and ensure user safety, recent works~\cite{zhao2025security, zhao20252} have adopted deep learning models for their strength in capturing complex user behavior, enabling personalized recommendation~\cite{xiao2023know, xiao2023user, wu2025align} and abnormal behavior detection~\cite{xiao2024make, zou2023iotbeholder}.
% While these systems offer significant convenience, their close integration with users' private lives also presents substantial risks to both security and privacy \cite{anderson2014synthetic}.
% To improve user convenience and safety in smart homes, recent research focuses on integrating intelligent models to detect harmful or abnormal behaviors~\cite{xiao2024make, xiao2023know}. They make recommendations on how to operate IoT actions based on users' historical behavior and contextual information, making this approach an effective solution paradigm for home IoT intelligence and security.

% However, existing smart home intelligent models are trained in a one-off manner using pre-collected datasets, which renders them either incapable of generalizing to diverse scenarios or dependent on substantial time and labor for data collection. 
% In the real world, user behaviors  are influenced by factors like season, lifestyle, and work status, leading to subtle and significant changes over time.  
However, these models are usually trained on static pre-collected datasets, which limits generalizability and requires costly data collection. In practice, user behaviors may evolve over time due to various reasons such as seasonal changes, lifestyle shifts, and work status variations, resulting in shifts in behavioral patterns. For example, users may use heaters frequently during cold seasons but turn to use cooling devices instead during warm seasons.
% Although pre-collected IoT datasets are derived from real-world scenarios, they capture only a small snapshot of a user's extended usage period. 
% As a result, they lack the ability to account for dynamic changes and may lose their effectiveness over time, causing significant issues when these models are applied in real-world settings. 
%As a result, actions once considered normal may later be considered as abnormal or even harmful, while behaviors previously flagged as anomalous may become innocent in a new context. 
%Such behavioral drifts can significantly increase downgrade the performance of intelligent models, e.g., actions once considered normal may now be considered as abnormal or even harmful. 
%For example, behavior considered normal in the original context may be deemed abnormal or harmful in a different scenario, while behavior previously classified as abnormal may become normal in a new context. This mismatch can result in significantly higher rates of missed detections and false alarms. % Another example of this limitation is the use of fixed patterns learned in a single, static scenario to predict user behavior in dynamic environments, which proves both inefficient and unreliable. 
Recent studies~\cite{rieger2023argus, xiao2023know} highlight that  models often struggles to adapt to such significant changes, resulting in degraded performance when facing unforeseen or changing user behaviors.

% leading to an excessively high false positive rate and an inability to handle unforeseen events.

% these shortcomings. 
% For instance, ARGUS~\cite{rieger2023argus} notes that the model struggles to adapt to significant changes, leading to an excessively high false positive rate and an inability to handle unforeseen events. Similarly, SmartUDI~\cite{xiao2023know} points out that such models lack adaptability to specific environments and face challenges in managing the complexities introduced by multiple users.
A common approach to address this problem is to continuously collect new data for model retraining. However, this method has two main limitations. First, gathering sufficient data often takes weeks or even months, during which the outdated model may suffer from degraded performance. Second, frequent collection of user behavior data raise significant privacy concerns, making this approach less suitable for deployment in privacy-sensitive environments like smart homes.
Latest large language models (LLMs) have demonstrated strong capabilities in semantic understanding, logical reasoning, and contextual generalization~\cite{gao2024chatiot}. These strengths make LLMs a promising tool for synthesizing user behavior sequences that reflect contextual changes—such as seasonal variations or lifestyle shifts—while aiming to maintain coherence with unaltered behavioral patterns observed in historical data.

%In this paper, we propose \schemename, an LLM-based semantic-aware user behavior sequence synthesis framework that generates behavior sequences in response to contextual variations, supporting the continual adaptation of deep learning models in smart home environments. 
However, developing an effective LLM-based synthesis framework face several challenges. 
First, user behavior sequences lack clear structural boundaries.
Unlike natural language, which is segmented by punctuation, user behavior data tends to be continuous and unstructured, making it difficult for LLMs to identify consistent behavioral patterns and underlying intentions.
Second, the limited context window of LLMs constrains the processing of large-scale behavior data.
Lengthy behavioral sequences with high token counts incur substantial computational costs and latency.
Third, historical data lacks explicit cues for user behavior regularities.
Without clear indicators of regularities (e.g., the user’s habitual tendencies), LLMs may struggle to generate sufficiently authentic and consistent behavior sequences.
Fourth, ensuring the quality and safety of synthetic sequences is non-trivial.
LLMs may produce implausible or incoherent behavior combinations that could negatively affect downstream model training.

\begin{figure}[ht]
\centering
\includegraphics[width = .4\textwidth]{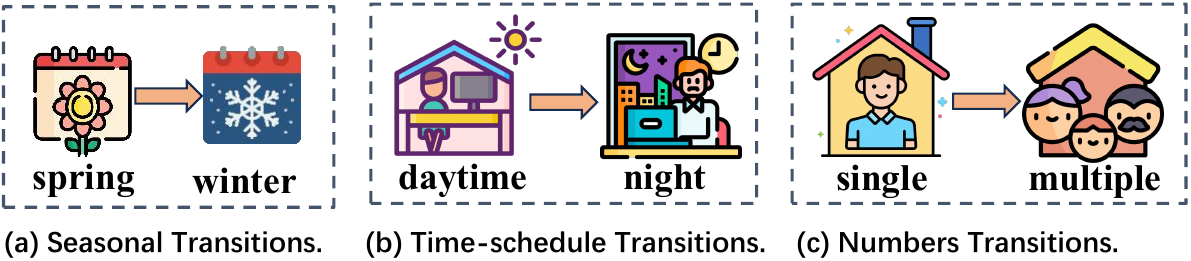}
\caption{Different type of environmental context transitions}
\label{fig:type}
\end{figure}
In this paper, we propose SmartGen, an LLM-based semantic-aware user behavior sequence synthesis framework that generates behavior sequences for contextual transitions, supporting the continual adaptation of intelligence models in smart homes. SmartGen address the aforementioned challenges by a four-stage design.
%to ensure the synthesis is semantically coherent, contextually diverse, computationally efficient, and reliable in quality.

First, we design a \textbf{Time and Semantic-aware Split (TSS)} module to divide long user behavior sequences into manageable and meaningful subsequences. TSS ensures semantic coherence by grouping related behaviors while applying time thresholds to limit gaps and segment duration.
%TSS jointly enforces semantic coherence and temporal constraints: it uses semantic checks to keep closely related behaviors together, while applying two time-based thresholds to prevent excessive gaps between actions and to limit the total duration of each segment. 
% Second, we propose S\textbf{emantic-aware Sequence Compression (SSC)}, which reduces input length by selecting structurally representative behavior sequences. Instead of relying on token-level redundancy, SSC uses a transformer-based autoencoder to embed behavior sequences and applies clustering on the embeddings to retain a compact yet informative subset of the data.
Second, we propose \textbf{Semantic-aware Sequence Compression (SSC)} to reduces input length by selecting semantically representative behavior sequences. SSC utilizes a transformer-based encoder to encode user behavior sequences, obtain representations in a semantic space, and applies clustering on the semantic representations to retain information-rich subsets of the original data.
% Third, we design a \textbf{Graph-guided Sequence Synthesis (GSS)} method to guide the LLM in generating behavior sequences that remain consistent with underlying patterns in the original data.
% GSS builds a behavior relationship graph from historical sequences, extracts frequent action transitions via a top-$k$ strategy, and encodes them into JSON prompts, enabling the LLM to preserve global behavioral regularities while adapting to contextual changes. By working jointly, SSC and GSS enable a synthesis process that achieves both compression efficiency and alignment with original behavior distributions—balancing local representativeness with global structural consistency.
Third, we design a \textbf{Graph-guided Sequence Synthesis (GSS)} method to guide the LLM to generate behavior sequences that remain consistent with underlying patterns in the original data.
GSS builds a behavior relationship graph from historical sequences, extracts frequent action transitions via a top-$k$ strategy, and encodes them into JSON prompts, enabling the LLM to preserve global behavioral regularities. SSC and GSS jointly ensures the synthesis process is both compression efficient and can alignment with original behavior distributions outside the target context transition.
%To address above challenges, we propose \schemename, a novel LLM-based framework for data synthesis in smart homes. First, we design a \textbf{\timesplit \  (\TS)} method to split long user behavior sequences into meaningful subsequences by applying two time thresholds that prevent excessive time intervals and cumulative durations, while preserving semantic coherence by keeping related behaviors together. Second, we propose a \textbf{\compression \ (\cp)} to reduce the token count. By utilizing an autoencoder to embed behavioral sequences, \cp \ assess the pattern information within the data sequence, subsequently selecting representative sequences by clustering for retention. Third, we propose a \textbf{\graph \ (\gp)} method to guide the data synthesis process. \gp \ first builds a behavior relationship graph from historical sequences, then uses this global information to guide the LLM in generating behaviors that differ from historical patterns, ensuring diversity in the synthetic dataset. 
Fourth, we propose a \textbf{\filter \ (\dof)} to remove noise and anomalies in the synthetic data. {\dof} first identifies outliers through reconstruction loss, then evaluates their utility to determine which data to discard.
% In combination with the \textbf{\cp} method, it achieves a perfect balance between data compression and information retention. 
Experiments on two smart home tasks show \schemename \ improve the model's adaptive performance by an average of 85.43\% and 70.51\%, respectively.

\begin{figure*}[ht]
\centering
\includegraphics[width = .90\textwidth]{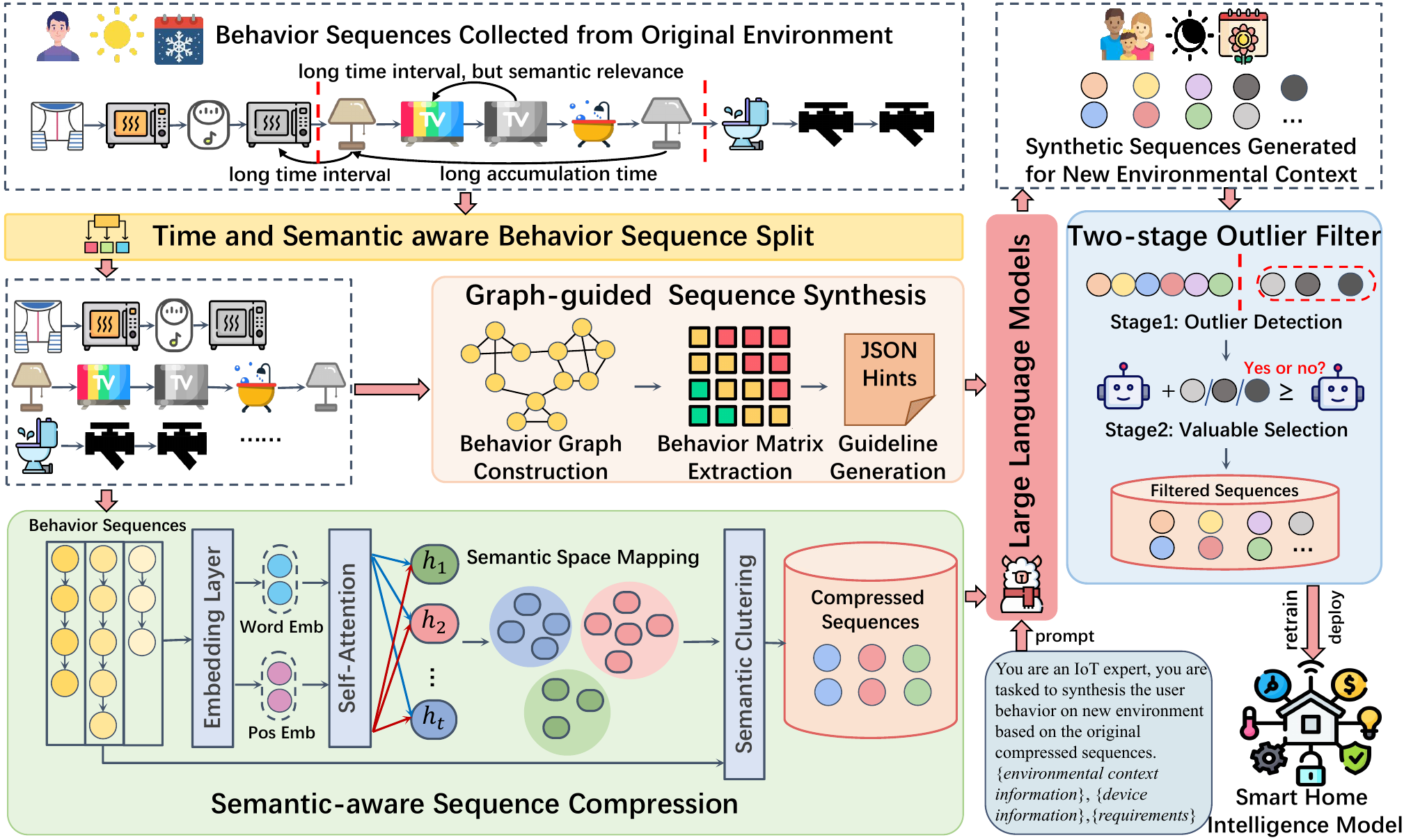}
\caption{Overview of \schemename.}
\label{fig:flow}
\end{figure*}

\section{Related Work}
\subsection{Behavior-Centric Smart Home Modeling}
% The increasing number and diversity of IoT devices has expanded the potential attack surface of many IoT systems, making them more vulnerable to security threats. Adversaries may use various technical means to induce abnormal behavior or disrupt normal execution of IoT processes, potentially causing economic losses or even physical harm to users in smart homes~\cite{delay1, delay2}. 
A variety of intelligent models have been developed to enable services in smart homes, such as anomaly detection and user behavior prediction~\cite{wang2023iot, wang2020iot, sikder2019aegis, fu2021hawatcher, gu2020iotgaze}. For instance, ARGUS \cite{rieger2023argus} employs a GRU-based autoencoder to detect contextual anomalies; SmartGuard \cite{xiao2024make} integrates multi-dimensional time encoding to handle temporal inconsistencies. For predictive tasks, IoTBeholder~\cite{zou2023iotbeholder} adopts an attention-based LSTM to capture user habits, while SmartSense~\cite{smartsense} introduces a context-aware encoder for modeling device-action correlations. More recent efforts like DeepUDI \cite{xiao2023user} and SmartUDI \cite{xiao2023know} utilize graph neural networks and contrastive learning to model user routines and intentions.

These models rely heavily on user behavior data for training and continual adaptation. However, collecting such data in practice is often difficult due to privacy concerns, deployment constraints, and variability across environments. This mismatch between data requirements and availability motivates the need for effective behavior data synthesis, which can serve as a foundation for training robust and generalizable smart home models.
% This paper aims to generalize the task model in the smart home scenario through large language model data synthesis, which requires the use of data synthesis, large language model and other technologies, and mainly focuses on the two types of smart home models: anomaly detection and behavior prediction. 
%The following will briefly introduce the related work in these three areas and explain their relationship with the method of this paper.

\subsection{Data Synthesis for IoT}
% Data synthesis is a widely explored topic in the field of machine learning and data science, enabling the generation of tabular data, images, text, audio, and more~\cite{datasynthesis}. 
% Training intelligent models for smart homes requires substantial user behavior data, which is often limited or hard to obtain in practice. This motivates the use of data synthesis to augment or simulate realistic behavior patterns.
To address data scarcity, several efforts have explored synthetic data generation for IoT. Anderson et al.\cite{anderson2014synthetic} propose synthesizing XML-based IoT data. SA-IoTDG\cite{mondal2022situation} uses hidden Markov models to simulate context-aware sensor streams. Other works~\cite{redvzovic2017ip, yin2022practical, patki2016synthetic} focus on generating network-level or packet-level sequences, such as IP traffic~\cite{redvzovic2017ip} using Markov chains or time-aligned flows~\cite{yin2022practical} based on 5-tuples. IoTGemini~\cite{li2024iotgemini} employs sequential GANs to generate high-fidelity traffic data.
However, most existing methods concentrate on simulating protocol-level or application-specific IoT data, and rarely consider the synthesis of human-centric user behavior~\cite{cheng2019pac, wang2020packetcgan, ring2019flow}. Moreover, they often lack adaptability to generate data for unseen scenarios or dynamically evolving contexts.

\subsection{Large Language Models}
% Large language models (LLMs)~\cite{gptseries, llamaseries} have achieved remarkable success across various tasks (e.g., coding, math).

% Beyond NLP, LLMs have demonstrated immense potential in other fields (e.g., coding \cite{xiao2024interaction2code}, math), offering transformative capabilities.
%but also introducing challenges, such as high inference costs.
Large language models (LLMs)\cite{gptseries, llamaseries, deepseekr1}, trained on vast and diverse corpora, possess exceptional semantic understanding, broad knowledge coverage, and strong instruction-following capabilities. They have demonstrated impressive performance across a wide range of open-ended tasks—such as code generation\cite{xiao2024interaction2code, xiao2025designbench, wan2024mrweb}, mathematics~\cite{azerbayev2023llemma, yu2025formalmath}, design~\cite{tang2025slidecoder} and computer use~\cite{chen2025survey, zou2025queryattack}—even in zero-shot~\cite{li2025open} or few-shot~\cite{bendou2025proker} scenarios. These capabilities align well with some of the key demands in smart home behavior modeling, which often requires implicit knowledge, logical reasoning, and generalization beyond limited behavioral observations.

However, leveraging LLMs in this setting presents several practical challenges~\cite{huang2024advancing}. First, their limited context window and the quadratic complexity of self-attention make it difficult to process long user behavior inputs, leading to high inference latency and memory consumption~\cite{chen20252}. Second, lengthy inputs may cause LLMs to lose focus on salient information—a phenomenon known as the “lost-in-the-middle” effect~\cite{liu2024lost}—which can undermine the plausibility and coherence of the generated behaviors.

\section{Problem Formulation}
\label{sec:pf}
In this section, we formally define the problem of context-aware behavior sequence synthesis in smart home environments.
Let $\mathcal{D}$ denote the IoT devices set, $\mathcal{A}$ denote the set of device actions set.

\noindent(\textbf{Behavior}) A behavior $b=(t, d, a)$ is a 3-tuple consisting of time stamp $t$, device $d \in \mathcal{D}$ and action $a \in \mathcal{A}$. For example, behavior \textit{b = (2022-08-04 18:30, air conditioner, air conditioner:switch on)} describes the behavior ``\textit{swich on the air conditioner}'' at 18:30 on 2022-08-04.

% \begin{myDef}   
% (Behavior Sequence) Given a set of behavior $b^{u}_{i}$ that occur in session $u \in \mathcal{U}$, the behavior sequence $s_{u}=[b^{u}_{1}, b^{u}_{2}, \cdots, b^{u}_{n}]$, where $b$ is ordered by timestamps.
% \end{myDef}

\noindent(\textbf{Behavior Sequence}) A behavior sequence $s=[b_{1}, b_{2}, \cdots, b_{n}]$ is an ordered list of behaviors arranged by timestamps, where $n$ is the length of the sequence.  The dataset, denoted as $\mathcal{S}$, comprises a collection of such behavior sequences.

\noindent(\textbf{Behavior Sequences Synthesis}) Input the original behavior sequences $\mathcal{S}$, the change in environmental context from $E_{\text{ori}}$ to $E_{\text{new}}$, device information $\mathcal{D}$, and prompt into a LLM to obtain synthesized behavior sequences $\mathcal{S}'$ that remain consistent with historical behavior patterns while adapting to the new environment $E_{\text{new}}$.

\noindent(\textbf{Smart home environmental context changes}) We consider three types of smart home environmental context changes, as illustrated in Figure~\ref{fig:type}\footnote{The proposed framework's LLM-powered design allows handling of other smart home context variations as well thanks to its inherent generalization capabilities.}: 1) Seasonal Transitions \textbf{(ST)} refers to the change of seasons, e.g., from winter to spring. In winter, users often use heaters frequently. In spring, users may use air conditioners or fans instead; 2) Time-schedule Transitions \textbf{(TT)} refers to changes in users' primary activity periods, e.g., when users transition from daytime to nighttime activities due to occupational changes; 3) Number Transitions \textbf{(NT)} refers to changes in occupant counts, e.g., from single-occupancy to multiple-occupancy, which can lead to significant changes in behavioral patterns and behavior density.

\section{Methodology}
\subsection{Solution Overview}

%We propose the \schemename \, a framework that uses a large language model to generate synthetic data for adaptive training of task models, in order to achieve flexible and adaptable smart home systems. 
% As illustrated in Fig.~\ref{fig:flow}, \schemename \ consists of five main modules: a {\timesplit} (\TS) module, a {\compression} (\cp) module, a {\graph} (\gp), an IoT Synthetic Data Generation module, and the last {\filter} (\dof) module. 
As illustrated in Fig.~\ref{fig:flow}, \schemename \ consists of four main modules: a {\timesplit} (\TS) module, a {\compression} (\cp) module, a {\graph} (\gp), and the last {\filter} (\dof) module. 

First, the behavior sequences collected from original environmental context are processed by the {\TS} module, which segments the long sequence using a two-level time intervals and semantic checker. The segmented sequence is then passed to the {\cp} module, where it is embedded through an embedding layer and encoded via an attention-based encoder. Within the semantic space map, the sequence is clustered and compressed, enabling the selection and retention of truly representative subsequences.

Concurrently, the {\gp} module captures the global frequency information from the sequences, constructs a user behavior transition graph and corresponding matrix, and generates a guideline that reflects user behavior habits. This helps mitigate the loss of global information during compression, thereby guiding the synthesis more aligned with both user characteristics and real scenarios.

Subsequently, the sequences and guideline are combined with COT-based and specially designed instructions, fed into the LLM for data synthesis adapted to the new environmental context. Finally, the {\dof} module filters out useless noise and potentially harmful sequences from the synthetic data to ensure its quality and reliability. The refined synthetic sequences are ultimately used to retrain the smart home model, which is then deployed within the smart home to enhance its adaptability to dynamic, real-world environments.

% The Structure Pattern Perception Compression module measures the richness of structural pattern information in IoT sequence data through an autoencoder and assigns importance scores to the sequences. By ranking these scores, the module filters and preserves truly representative IoT sequence data, thereby enhancing the efficiency of data generation. The IoT Synthetic Data Generation module takes the IoT sequence data from the Structure Pattern Perception Compression module as input, adds corresponding instructions and dictionaries, and generate IoT sequence data for new scenes. The generated data can then assist task models in adaptive training to improve the generalization and continuous reliability of the models.

\subsection{Time and Semantic-aware Split}

% First, to uncover users’ potential behavior patterns and habits from IoT data and to mitigate the problems caused by long behavior sequences, we introduce Time and Semantic Aware Split (TSAS) method to address C1. This approach splits excessively long behavior sequences by setting two levels of time-span thresholds. It ensures that the time span between individual behaviors within the split subsequences remains within a reasonable range, while also limiting the total duration of each subsequence. This strategy aims to prevent the potential masking or loss of the semantic meaning and user intent behind the behavior, thereby simplifying the model’s complexity in deeply processing user behavior sequences. 

% In reality, the raw data of user behaviors collected are long and continuous sequences. Since behavior sequences do not have obvious interval symbols like natural languages, we need to design a segmentation algorithm to perform segmentation.
Overly long temporal sequences may involve multiple behavioral patterns, increasing their complexity and posing challenges for large language models to interpret effectively. As real-world smart home behavior data is recorded as uninterrupted sequences of timestamped actions without syntactic signals for segmentation, a dedicated strategy is needed to appropriately split the sequences based on semantic and temporal features, which should satisfy three requirements: 1) preventing behaviors with excessively long time intervals from appearing within a single sequence; 2) ensuring the total duration of each sequence remains within reasonable bounds; 3) maintaining semantic coherence by avoiding the separation of semantically related behaviors across different sequences.

To meet above requirements, we design the \textit{Time and Semantic aware Split} (TSS) algorithm as shown in Algorithm~\ref{alg:tsas}.The algorithm first evaluates semantic coherence (Line 6) using the SemanticChecker function to avoid splitting semantically related behaviors (e.g., the operation of turning on and off a water valve should not be divided into two sequences). When the time interval between consecutive behaviors exceeds the maximum threshold $\Delta t_{max}$ (Line 8), the algorithm triggers segmentation to prevent behaviors separated by excessive time gaps from being grouped together. Additionally, the algorithm monitors the total sequence duration and triggers segmentation when the current sequence duration exceeds $T_{max}$ (Line 12) to maintain manageable sequence lengths.

% ensure behaviors within the same sequence maintain semantic relatedness; if semantic coherence is violated, it proceeds to evaluate temporal constraints. When the time interval between consecutive behaviors exceeds the maximum threshold $\Delta t_{max}$ (Line 8), the algorithm triggers segmentation to prevent behaviors separated by excessive time gaps from being grouped together. Additionally, the algorithm monitors the total sequence duration and triggers segmentation when the current sequence duration exceeds $T_max$ (Line 12) to maintain manageable sequence lengths. 

% This hierarchical decision structure prioritizes maintaining meaningful behavioral contexts while respecting temporal boundaries, ensuring that the output sequences are both semantically coherent and temporally constrained.

\begin{algorithm}[h]
\caption{Time and Semantic Aware Split Algorithm}
\label{alg:tsas}
\begin{algorithmic}[1]
\REQUIRE Raw behavior sequence $S_{raw} = [b_1, b_2, \ldots, b_m]$, maximum time interval $\Delta t_{max}$, maximum sequence duration $T_{max}$
% semantic similarity threshold $\theta$
\ENSURE Set of segmented behavior sequences $\mathcal{S} = \{s_1, s_2, \ldots, s_k\}$

\STATE Initialize $\mathcal{S} \leftarrow \emptyset$, $s_{current} \leftarrow [b_1]$, $t_{start} \leftarrow b_1.t$
\STATE $i \leftarrow 2$

\WHILE{$i \leq m$}
    \STATE $b_{curr} \leftarrow b_i$
    \STATE $b_{prev} \leftarrow s_{current}[-1]$ 
    % \COMMENT{Last behavior in current sequence}
    % \STATE \COMMENT{Check time interval constraint}
    % \IF{$\text{SemanticChecker}(s_{current}, b_{curr}) < \theta$}
    \IF{$\text{SemanticChecker}(s_{current}, b_{curr})$}
        \STATE continue
    \ELSIF{$b_{curr}.t - b_{prev}.t > \Delta t_{max}$}
        \STATE $\mathcal{S} \leftarrow \mathcal{S} \cup \{s_{current}\}$
        \STATE $s_{current} \leftarrow [b_{curr}]$
        \STATE $t_{start} \leftarrow b_{curr}.t$
    \ELSIF{$b_{curr}.t - t_{start} > T_{max}$} 
    
    % \COMMENT{Check sequence duration constraint}
    
        \STATE $\mathcal{S} \leftarrow \mathcal{S} \cup \{s_{current}\}$
        
        \STATE $s_{current} \leftarrow [b_{curr}]$

        \STATE $t_{start} \leftarrow b_{curr}.t$
    
    % \ELSIF{$\text{SemanticSimilarity}(s_{current}, b_{curr}) < \theta$} 
    
    % % \COMMENT{Check semantic coherence}
    %     \STATE $\mathcal{S} \leftarrow \mathcal{S} \cup \{s_{current}\}$
        
    %     \STATE $s_{current} \leftarrow [b_{curr}]$, $t_{start} \leftarrow b_{curr}.t$
    \ELSE
        \STATE $s_{current} \leftarrow s_{current} \cup \{b_{curr}\}$ 
        % \COMMENT{Add behavior to current sequence}
    \ENDIF
    
    \STATE $i \leftarrow i + 1$
\ENDWHILE

\STATE $\mathcal{S} \leftarrow \mathcal{S} \cup \{s_{current}\}$ 
%\COMMENT{Add the last sequence}

\RETURN $\mathcal{S}$

\end{algorithmic}
\end{algorithm}

% \noindent\textbf{Function SemanticSimilarity:}
% \begin{algorithmic}[1]
% \FUNCTION{SemanticSimilarity}{$s, b$}
%     \STATE $devices_{s} \leftarrow \{b_i.d \mid b_i \in s\}$ \COMMENT{Extract devices in sequence $s$}
%     \STATE $controls_{s} \leftarrow \{b_i.c \mid b_i \in s\}$ \COMMENT{Extract controls in sequence $s$}
    
%     % \STATE $sim_{device} \leftarrow \mathbf{1}_{b.d \in devices_{s}}$ \COMMENT{Device overlap indicator}
%     \STATE $sim_{control} \leftarrow \max_{c \in controls_{s}} \text{ControlSimilarity}(b.c, c)$ \COMMENT{Control similarity}
    
%     \STATE $similarity \leftarrow \alpha \cdot sim_{device} + (1-\alpha) \cdot sim_{control}$
%     \RETURN $similarity$
% \ENDFUNCTION
% \end{algorithmic}

\subsection{Semantic-aware Sequence Compression}
% In smart home scenarios, users frequently interact with IoT devices. For example, the SmartSense~\cite{smartsense} dataset contains 2,000 FR data entries and 8,000 US data entries. After converting these entries to text, the resulting data spans 914k to 3,657k tokens. Such extensive text contains way too much redundancy when used as prompt input. When combined with external knowledge, documents, or dictionaries, the prompt length often grows to an unmanageable extent. Research indicates that excessively long input texts can cause large language models (LLMs) to forget portions of the intermediate content, thereby reducing the overall inference performance. Additionally, overly lengthy texts result in high computational costs and increased inference latency.
% This poses two key challenges: first, excessively long prompts can significantly slow down the response speed of large language models (LLMs) and may even exceed their processing limits. Second, longer prompts lead to higher LLM call costs. 

% In fact, in many cases, the datasets itself contain significant information redundancy, which provides the feasibility for prompt compression.

In practice, the datasets often contain significant redundancy in information. %Research indicates that excessively long input texts can cause large language models (LLMs) to forget portions of the intermediate content, thereby reducing the overall inference performance. Additionally, overly lengthy texts result in high computational costs and increased inference latency.
Overly long inputs can cause LLMs to forget intermediate content, hurting inference performance, while also increasing computational costs and latency.

% Therefore, various prompt compression methods have been proposed. In LLMLingua, information entropy and conditional probability are employed to select the tokens from the text that most contribute to the accurate generation of an answer. ChatIoT, on the other hand, reduces input by eliminating irrelevant information through similarity calculations. However, IoT data exhibits distinct characteristics. Firstly, IoT data primarily consists of automated device actions and user behaviors. While device actions exhibit a certain level of spontaneity, user behaviors are influenced by the user’s intentions, characterized by short-term continuity but long-term variability. As a result, IoT data does not have the continuous contextual relations and semantic coherence typical of pure text content. This leads to weak inter-sequence correlations in the dataset, making it difficult to apply conditional probability for compression. Secondly, in IoT data, redundancy cannot be effectively measured using text or item similarity between sequences, which risks mistakenly deleting unique combinations of actions. 

To address these, the Semantic-aware Sequence Compression ({\cp}) module is proposed to compress the datasets by leveraging the overall structural information of sequences rather than only relying on text-based information.

First, we use a transformer-based autoencoder to fully learn the original behavior sequences dataset $\mathcal{S}$:
\begin{equation}
  \theta^*=\arg \min _\theta \textstyle\sum_{i=1}^N L\left(\operatorname{Autoencoder}\left(\mathcal{S}^{(i)}, \theta\right), \mathcal{S}^{(i)}\right),
\end{equation}
where $\theta^*$ represents the optimal parameters of the model.
Each behavior sequence $s=[b_{1}, b_{2}, \cdots, b_{t}]$, where $b_{t}$ is the behavior vector of the $t$-th time step, is mapped to a vector space of fixed dimension through word embedding and position encoding:
\begin{equation}
    \mathbf{P E}(t)=\left[\sin \left(\frac{t}{10000^{2 i / d}}\right), \cos \left(\frac{t}{10000^{2 i / d}}\right)\right]_{i=1}^{d / 2},
\end{equation}
where $t$ is the position information of the word, $i$ is the dimension index of the position encoding, and $d$ is the dimension of the embedding space. In this way, the input vector $s$ is incorporated into the embedding $\mathbf{h}$:
\begin{equation}
    e_t=\mathbf{E m b}\left(b_t\right)+\mathbf{P E}(t),
\end{equation}
\begin{equation}
    \mathbf{h} = [e_{1}, e_{2}, \cdots, e_{t}],
\end{equation}
where $\mathbf{E m b}\left(b_t\right)$ is word embedding, $\mathbf{P E}(t)$ is position embedding.

Then, $\mathbf{h}$ is fed into the encoder of transformer, which is composed of multiple self-attention layers and feed-forward networks. The operation steps of each layer of the encoder can be expressed as:
\begin{equation}
\mathrm{Q}=\mathbf{h}\mathrm{W}^Q_{\theta^*}, \mathrm{~K}=\mathbf{h}\mathrm{W}^K_{\theta^*}, \mathrm{~V}=\mathbf{h}\mathrm{W}^V_{\theta^*}, 
\end{equation}
where $\mathrm{W}^Q_{\theta^*}, \mathrm{W}^K_{\theta^*}, \mathrm{W}^V_{\theta^*}$ are the trained transformation matrices. The attention score $\mathbf{A}$ is computed by:
\begin{equation}
\mathbf{A} = \operatorname{Attention}(Q, K, V)=\operatorname{softmax}\left(\frac{Q K^T}{\sqrt{d_k}}\right) V,
\end{equation}
where $d_{k}$ is the dimension of $K$. Multi-head attention is applied to improve the stability of the learning process and achieve higher performance. Then, the position-wise feedforward neural network (FNN) and residual connections are adopted:
\begin{equation}
\operatorname{FFN}(\mathbf{A})=\operatorname{ReLU}\left(\mathbf{A} \mathbf{W}^{\theta^*}_1+\mathbf{b}^{\theta^*}_1\right) \mathbf{W}^{\theta^*}_2+\mathbf{b}^{\theta^*}_2,
\end{equation}
where $\mathbf{W}^{\theta^*}_1$, $\mathbf{W}^{\theta^*}_2$ are the weight matrices of the trained feedforward neural network, and $\mathbf{b}^{\theta^*}_1$, $\mathbf{b}^{\theta^*}_2$ are bias terms.

Finally, we get the representation $\mathbf{z}$ of $s$ in the semantic space:
\begin{equation}
\mathbf{z}=\operatorname{LayerNorm}\left(\mathbf{A}+\operatorname{FFN}\left(\mathbf{A}\right)\right)
\end{equation}
where $\mathbf{z}$ is the final output of the encoder, which includes the update of the self-attention mechanism and the feedforward neural network, representing the semantic characteristics of $s$. $\operatorname{LayerNorm}(\cdot)$ is the residual connection.

To evaluate the semantic relationship between two behavior sequences $s_i$ and $s_j$, we first obtain their semantic representations $\mathbf{z}_i$ and $\mathbf{z}_j$ using the transformer encoder described above. Then, we compute their cosine similarity as follows:
\begin{equation}
    c_{i, j}=\cos \left(\mathbf{z}_i, \mathbf{z}_j\right)=\frac{\mathbf{z}_i \cdot \mathbf{z}_j}{\left\|\mathbf{z}_i\right\|\left\|\mathbf{z}_j\right\|}.
\end{equation}

We set a compression threshold $\alpha$ and initialize an empty set $\mathbf{R}$ of indices to be removed. For each sequence $s_i$ not already marked for removal, we identify all subsequent sequences $s_j$ that are semantically similar (i.e., $c_{i,j} > \alpha$) and mark them for removal:
\begin{equation}
    \mathbf{R} \leftarrow \mathbf{R} \cup \left\{ j \mid j > i \land c_{i,j} > \alpha \right\}, \quad \text{for each } i \notin \mathbf{R}
\end{equation}
\begin{equation}
    \mathbf{U} = \left\{ s_i \mid i \notin \mathbf{R} \right\}
\end{equation}
Where $\mathbf{U}$ represents the compressed sequences data.

This process performs a lightweight form of semantic clustering by retaining only one representative from each set of highly similar sequences, effectively reducing redundancy while preserving the semantic diversity of the dataset.

\subsection{Graph-guided Sequence Synthesis}
User behavior pattern information contained in IoT data can be categorized into two dimensions: diversity and frequency. While the data compressed by {\cp} retains the diversity dimension—a broad range of distinct behavior combinations and interaction patterns—it inevitably loses frequency information, which reflects how often specific behaviors occur and represents the user's habitual tendencies. Preserving frequency information is essential for ensuring that the synthesized data reflects the consistency of user behaviors. To address this, we propose {\gp} to recover the lost frequency information and reduce semantic loss introduced during {\cp}.
%The user information contained in IoT data can be classified into user behavior pattern information and user behavior habit information. {\cp} successfully eliminates highly repetitive behavior sequences from IoT data, thereby achieving data compression. However, while the compressed data retains various user behavior combinations and patterns effectively, it inevitably loses frequency information of user behaviors, leading to the loss of frequent behavior data, i.e., user behavior habits. To preserve the record of user behavior habits and minimize information loss caused by data compression, {\gp} is proposed. 
{\gp} consists of three components: a global behavior information graph, an action transition matrix, and JSON format-guided hints.

% \begin{figure*}[ht]
%     \subfigure[The example of global behavior information graph.]{
%     \label{graph1}
%     \centering
%     \includegraphics[width = .6\textwidth]{figure/graph_1.pdf}
%     }
%     \subfigure[The example of action transition matrix.]{
%     \label{graph2}
%     \centering
%     \includegraphics[width = .35\textwidth]{figure/graph_2.pdf}
%     }
%     \caption{The implementation process of {\gp}.}
%     \label{Graph}
% \end{figure*}

\begin{figure}[ht]
    \centering
    \includegraphics[width = .45\textwidth]{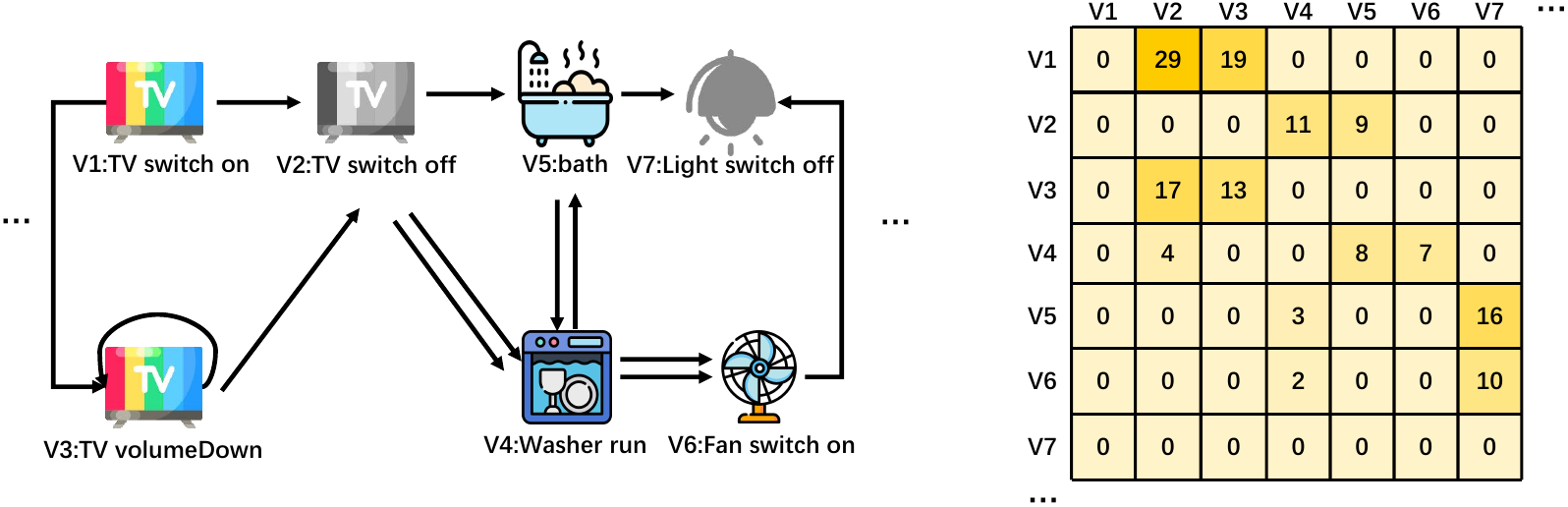}
    \caption{An example of {\gp}.}
    \label{Graph}
\end{figure}

%\subsubsection{Global behavior information graph}
% In the IoT dataset of size $N$, for each sequence of user behavior $S_q$, it consists of $n$ quadruples:
% \begin{equation}
%   S_q = [D_0,\ T_0, \ I_0, \ A_0, \ D_1,\ T_1, \ I_1, \ A_1, \ D_2,\ T_2, \ I_2, \ A_2, \ ... \ D_n,\ T_n, \ I_n, \ A_n],
% \end{equation}
% among them, $D_i$ represents the day of the week when the behavior occurs, $T_i$ represents the timestamp of the behavior occurring, specific to the hour, $I_i$ represents the IoT device that the user interacts with, and $A_i$ represents the action performed by the device.
% The $Device\_actions_q$ of the device action transition sequence was extracted from $S_q$:
% \begin{equation}
%   Device\_actions_q = [A_0, \ A_1, \ A_2, \ ... \  A_n] = A_0\xrightarrow{}A_1\xrightarrow{}A_2\xrightarrow{}...\xrightarrow{}A_n
% \end{equation}

The global behavior information graph $G = (V, E)$ is a directed graph, where $V$ denotes the set of distinct user actions and $E$ the set of directed edges representing observed transitions between actions. To construct $G$, we sequentially traverse each behavior sequence in the original dataset. For every pair of consecutive actions $(V_i, V_{i+1})$, if the edge $(V_i, V_{i+1})$ does not exist in $E$, we add it to $E$ and initialize its weight to 1. Otherwise, we increment its existing weight. The edge weight $w(V_i \rightarrow V_{i+1})$ is updated as:

\begin{equation}
w(V_i \rightarrow V_{i+1}) =
\begin{cases}
1, & \text{if } (V_i, V_{i+1}) \notin E  \\
w(V_i \rightarrow V_{i+1}) + 1, & \text{if } (V_i, V_{i+1}) \in E.
\end{cases}
\end{equation}

% The global graph $G$ is a directed graph. The edge set is denoted as $E$. The weight of each edge $w(A_{i} \xrightarrow{} A_{i+1})$ records the number of transitions from $A_{i}$ to $A_{i+1}$ and is calculated as follows:
% \begin{equation}
% w(A_{i} \xrightarrow{} A_{i+1}) =
% \begin{cases}
% 0, & \text{if } (A_{i}, A_{i+1}) \notin E \\
% w(A_{i} \xrightarrow{} A_{i+1}) + 1, & \text{if } (A_{i}, A_{i+1}) \in E
% \end{cases}
% \end{equation}

%It is worth noting that in the experimental setup of this paper, the graph $G$ is derived from the original dataset of each scenario, and the relevant information about the test dataset is not recorded to avoid possible data leakage.

%\subsubsection{Action transition matrix}
We extract an action transition matrix $M \in \mathbb{N}^{|V| \times |V|}$ from the behavior graph $G = (V, E)$. Each entry $M_{i,j}$ records the number of times action $V_j$ directly follows action $V_i$ in the original dataset:
\begin{equation}
M_{i,j} =
\begin{cases}
w(V_i \rightarrow V_j) & \text{if } (V_i, V_j) \in E \\
0 & \text{otherwise}.
\end{cases}
\end{equation}
This matrix captures the empirical transition counts between user behaviors and serves as a statistical prior for guiding synthesis.

Then, we extract high-frequency user behavior transitions directly from the action transition matrix $M$, and format them into JSON-based textual hints.
For each row $i$ in $M$, which represents the current action $V_i$, we retrieve all non-zero entries $M_{i,j}$ indicating transitions to subsequent actions $V_j$ along with their frequencies. We apply a top-$k$ selection to retain the $k$ most frequent transitions.
% \begin{equation}
% \begin{cases}
% E' = Sort(E,\ key=w)  \\
% E_{topk} = \{E'_1,E'_2,E'_3,...,E'_k\}
% \end{cases}
% \end{equation}

The selected top-$k$ transitions are converted into textual form and saved in a JSON file as generation hints, guiding the LLM to produce behavior sequences that reflect common user patterns.

\begin{table*}[ht]
\small
\setlength{\tabcolsep}{0.1em}
\caption{\large System message design for \schemename.}
\label{tab:findings}
\begin{tabular*}{\linewidth}{@{}l|l@{}}
\toprule
Elements                          & Content                                                                                                                                                                                                                                                                                                                                                                                           \\ \midrule
\multirow{1}{*}{Role}  & \begin{tabular}[c]{@{}l@{}}You're an IoT expert. You are very knowledgeable about user behavior and habits in smart homes. The user would like to ask you about the \\ possible changes in user behavior sequence after the change of smart home user habits and env.
\end{tabular}  \\ \cmidrule(r){1-2}

\begin{tabular}[c]{@{}l@{}}Task \\ Background \end{tabular}      
&\begin{tabular}[c]{@{}l@{}}The user will provide you with the user's previous life environment and the changed environment, the user's previous behavior sequence, \\ and a devices set and device states. And the user hope that you can use these devices and states to generate possible user behavior \\ sequences after the env changes based on the original user behavior sequence.\end{tabular} \\\cmidrule(r){1-2}

\begin{tabular}[c]{@{}l@{}}CoT Task \\ Definition \end{tabular}      
&\begin{tabular}[c]{@{}l@{}}Your task: First, select the possible new device states from the set of devices and device states which are also possible new user behaviors. \\
The second step is to reasonably add possible new user behaviors to the original user behavior sequences. The third step is to reasonably \\ continue and expand the sequence based on user behavior habits.\end{tabular} \\\cmidrule(r){1-2}

        \multirow{9}{*}{Requirements} & \begin{tabular}[c]{@{}l@{}}Please consider the devices that will be used in the new environment as widely as possible based on the set of devices. 
        \end{tabular} \\\cmidrule(r){2-2}
        & \begin{tabular}[c]{@{}l@{}}
         Please strictly follow the correspondence between the devices and device states to generate. Do not generate device states that do not \\ match the device. 
         \end{tabular} \\\cmidrule(r){2-2}
        & \begin{tabular}[c]{@{}l@{}}
        Please add as many new devices and device behaviors as possible to better adapt to changes in the environment.
        \end{tabular} \\\cmidrule(r){2-2}
        & \begin{tabular}[c]{@{}l@{}}
        Please make sure that the generated sequence is not a single behavior, but a sequence of consecutive behaviors.
        \end{tabular} \\\cmidrule(r){2-2}
        & \begin{tabular}[c]{@{}l@{}}
        Please also generate reasonable behavior time when generating, not just a single behavior.
        \end{tabular} \\\cmidrule(r){2-2}
        & \begin{tabular}[c]{@{}l@{}}
        The final generated behavior sequences set is in the format of <seq [['...'], ['...'], ['...']] seq>.
        \end{tabular} \\\cmidrule(r){1-2}
        % & \begin{tabular}[c]{@{}l@{}}
        % Please make modifications in the original sequence. The generated new behavior sequence set is also in the format \\ of [[...], [...], ...].
        % \end{tabular} \\\cmidrule(r){1-2}
        
\multirow{2.5}{*}{\begin{tabular}[c]{@{}l@{}}Scene \\ Information \end{tabular}} & 
        \begin{tabular}[c]{@{}l@{}}The original environment is {$E_\text{ori}$}.
        \end{tabular}  \\ \cmidrule(r){2-2}
        & \begin{tabular}[c]{@{}l@{}}The new environment is {$E_\text{new}$}. 
        \end{tabular}  \\ \cmidrule(r){1-2}

\multirow{2.5}{*}{\begin{tabular}[c]{@{}l@{}}Data \\ Information \end{tabular}} & 
        \begin{tabular}[c]{@{}l@{}}The user's previous sequence of behavior: \{user sequences $\mathcal{S}$\}.
        \end{tabular}  \\ \cmidrule(r){2-2}
        & \begin{tabular}[c]{@{}l@{}}The set of the possible device and device states: \{device information $\mathcal{D}$ \& $\mathcal{A}$\}. 
        \end{tabular}  \\ 
        \bottomrule
\end{tabular*}
\end{table*}

\subsection{Two-stage Outlier Filter}

To further enhance the reliability of the synthesized data, we propose a two-stage outlier filter that identifies and eliminates noise and semantically inconsistent samples from the synthesized data.

% we still need to ensure the security and reliability of synthetic data. In that case, {\dof} proposes an automated method to accurately and reliably remove noise and potential anomalies from synthetic data.

\subsubsection{Stage 1: Reconstruction Loss-based Outlier Detection}
% As shown in section~\ref{ablation}, considering that unfiltered synthetic data can already perform well on downstream tasks, we can assume that most of the synthetic data are normal and qualified data. And for possible noise data and abnormal data caused by a combination of erroneous behaviors, they are usually quite different from the normal data group. 
We first identify suspected noise and abnormal data based on reconstruction loss. For the synthetic sequences data $\mathcal{S}'$, we train a transformer-based autoencoder to learn the user behavior sequence semantics. The training objective minimizes the negative log-likelihood loss between the reconstructed output and the original input:
% we put it into the $\operatorname{Transformer}_{seq2seq}$ model for training:
\begin{equation}
    L=-\textstyle\sum_{i=1}^N y_i \log \left(p_i\right),
\end{equation}
\begin{equation}
  \theta^*=\arg \min _\theta\textstyle \sum_{i=1}^N L\left(\operatorname{Autoencoder}\left(\mathcal{S}'^{(i)}, \theta\right), \mathcal{S}'^{(i)}\right),
\end{equation}
where $\theta^*$ represents the optimal parameters of the autoencoder, and $N$ is the total number of sequences in $\mathcal{S}'$.

Once the model is trained, we use the optimized autoencoder to compute the reconstruction loss for each synthetic sequence $s'_i$:

\begin{equation}
\ell_{s'_i} = \mathcal{L}\left(\operatorname{Autoencoder}{\theta^*}(s'_i),\ s'_i\right),\quad s'_i \in \mathcal{S}',
\end{equation}
and aggregate them into a reconstruction loss set:
\begin{equation}
\mathcal{L}{\text{rec}} = \left\{ \ell{s'_1}, \ell{s'_2}, \ldots, \ell{s'_N} \right\}.
\end{equation}

A higher reconstruction loss $\ell_{s'_i}$ suggests that the model has difficulty learning the pattern of $s'_i$, which may indicate semantic inconsistency or noise in the generated data.

% \begin{equation}
%   \operatorname{Model}_{\mathcal{S}^{\prime}} = \operatorname{Autoencoder}_{\theta^*}(\mathcal{S}^{\prime}),
% \end{equation}
% where $\theta^*$ represents the optimal parameters of the model and $\operatorname{Model}_{\mathcal{S}'}$ represents the optimal model obtained by training with the dataset $\mathcal{S}'$ as input and target. $N$ stands for the number of sequences in the synthetic data. Then the reconstruction loss $\mathcal{L}_{rec}$ corresponding to each synthetic sequence $s'$ is obtained:
% \begin{equation}
%     \mathcal{L}_{rec}=\mathcal{L}\left(\operatorname{Model}_{\mathcal{S}^{\prime}}, \mathcal{S}^{\prime}\right),
% \end{equation}
% \begin{equation}
%     \mathcal{L}_{\text {rec}} = \left\{\ell_{s'_1}, \ell_{s'_2}, \ldots, \ell_{s'_n}\right\}, {s'_n} \in \mathcal{S'},
% \end{equation}
% A higher reconstruction loss indicates that the model trained on set $\mathcal{S}^{\prime}$ struggles to learn the behavioral patterns effectively, suggesting the presence of noise or anomalous behavior in the sequence.
To identify anomalous sequences, we analyze the distribution of reconstruction losses $\mathcal{L}_{\text{rec}}$ using the interquartile range (IQR) method. Specifically, we compute the 25th and 75th percentiles of the loss distribution as $Q_1$ and $Q_3$, respectively:
% \begin{align}
%  Q_1 &=percentile (\mathcal{L}_{rec}, 25), \\
%  Q_3 &=percentile (\mathcal{L}_{rec}, 75), \\
%  \mathrm{IQR} &=Q_3-Q_1, \\
%  U&=Q_3+1.5 \cdot \mathrm{IQR}, \\
%  \mathcal{S}_{non-out}&=\left\{s_i \mid \ell_{s'_i} \in \mathcal{L}_{rec},  \ell_{s'_i} \leq U\right\}, \\
%  \mathcal{S}_{out}&=\left\{s_i \mid \ell_{s'_i} \in \mathcal{L}_{rec}, \ell_{s'_i}>U\right\}, \\
% \end{align}
\begin{equation}
    Q_1 =percentile (\mathcal{L}_{\text{rec}}, 25), \ Q_3 =percentile (\mathcal{L}_{{\text{rec}}}, 75).
\end{equation}
We then calculate the interquartile range (IQR) as:
\begin{equation}
\mathrm{IQR} = Q_3 - Q_1.
\end{equation}
Based on this, we define the upper bound $U$ for non-outliers as:
\begin{equation}
U = Q_3 + 1.5 \cdot \mathrm{IQR}.
\end{equation}

Using this threshold, we divide the synthetic sequences into two sets: the non-outlier set $\mathcal{S}_{\text{non-out}}$, which includes sequences with reconstruction loss less than or equal to $U$, and the outlier set $\mathcal{S}_{\text{out}}$, which includes sequences with higher loss:
\begin{equation}
    \mathcal{S}_{\text{non-out}}=\left\{s_i \mid \ell_{s'_i} \in \mathcal{L}_{\text{rec}},  \ell_{s'_i} \leq U\right\},
    \mathcal{S}_{\text{out}}=\left\{s_i \mid \ell_{s'_i} \in \mathcal{L}_{\text{rec}}, \ell_{s'_i}>U\right\}.
\end{equation}
This statistical filtering step enables us to remove sequences that deviate significantly from the learned distribution, thereby improving the overall reliability of the synthetic data.

\subsubsection{Stage 2: Outlier Evaluation and Selection}
While Stage 1 identifies potential outliers based on reconstruction loss, some low-frequency but semantically correct behavior combinations may also be misclassified as outliers due to the fully compressed nature of the synthetic dataset. To avoid discarding such valuable sequences, we further assess the utility of each outlier. 

We first split the non-outlier set $\mathcal{S}_{\text{non-out}}$ into a training set $\mathcal{S}_{\text{train}}$ and a test set $\mathcal{S}_{\text{test}}$ in an 8:2 ratio, and compute a baseline reconstruction loss:
\begin{equation}
  \mathcal{L}_{\text{st}} = \operatorname{Mean}(\mathcal{L}\left(\operatorname{Autoencoder}_{\theta^*}(\mathcal{S}_{\text{train}}), \ \mathcal{S}_{\text{test}}\right)).
\end{equation}

Then, for each outlier sequence $\mathcal{S}_{{\text{out}}_i}$, we evaluate its utility by adding it to the training set and comparing the resulting test loss with the baseline. If including $\mathcal{S}_{{\text{out}}_i}$ leads to a lower or equal test loss, it is retained; otherwise, it is discarded:
\begin{equation}
\begin{aligned}
\text{Decision}(\mathcal{S}_{\text{out}_i}) = 
\begin{cases}
\text{Retain}, & \text{if } \mathcal{L}\left(\mathcal{M}(\mathcal{S}_{\text{train}} \cup \mathcal{S}_{\text{out}_i}),\ 
\mathcal{S}_{\text{test}}\right) \leq \mathcal{L}_{\text{st}} \\
\text{Delete}, & \text{otherwise}
\end{cases}
\end{aligned}.
\end{equation}

\section{Experiments}
% In this section, we conduct comprehensive experiments on three real-world datasets and two downstream tasks, i.e., Anomaly Detection (AD) and Behavior Prediction (BP). We also conduct ablation study, parameter study (shown in Appendix~\ref{para}), and case study to demonstrate the stability and interpretability of \schemename.
In this section, we conduct comprehensive experiments on three real-world datasets and two downstream tasks, i.e., Anomaly Detection (AD) and Behavior Prediction (BP). We also conduct ablation study, parameter study (shown in Appendix~\ref{para}), and case study to demonstrate the stability and interpretability of \schemename.
%answer the following research questions:
% \begin{itemize}
% \item RQ1. \textbf{Performance.} Compared with other methods, can \schemename \ achieve better performance when facing environmental context changing?
% \item RQ2. \textbf{Ablation study.} How will model performance change if we remove key modules of \schemename?
% \item RQ3. \textbf{Parameter study.} How do key parameters affect the performance of \schemename?
% \item RQ4. \textbf{Interpretability study.} Can \schemename \ give reasonable explanations for the results of generation?
% \item RQ5. \textbf{Action matrix analysis.} Does synthetic data exhibit a stronger alignment with the target domain than the original data?
%\end{itemize}

\begin{table*}[h]
\caption{Performance comparison in three real-world datasets for anomaly detection.}
\label{tab:comad}
\setlength{\tabcolsep}{0.5em}
% \resizebox{\textwidth}{60mm}{
\begin{tabular}{@{}ccccccccccc@{}}
\toprule
Dataset             & Type                & Metric   & \multicolumn{1}{c}{LOF} & \multicolumn{1}{c}{IF} & \multicolumn{1}{c}{Aegis} & \multicolumn{1}{c}{OCSVM} & \multicolumn{1}{c}{Autoencoder} & \multicolumn{1}{c}{ARGUS} & \multicolumn{1}{c}{TransAE} & \multicolumn{1}{c}{\schemename} \\ \midrule
\multirow{10}{*}{FR} & \multirow{3}{*}{ST} & Recall   & 0.0541                  & 0.8559                                                    & 0.3063                  & 0.4144                          &0.9910               & 0.9279                   & \textbf{1.0000}       & \underline{0.9886}                  \\
                    &                     & Precision & 0.0769                  & 0.3878                                                  & 0.1423                    & 0.3129                          & 0.4977       & 0.4928  & \underline{0.5000} &  \textbf{0.7632}                     \\ 
                    &                     & F1 Score & 0.0635                  & 0.5337                                                 & 0.1943                    & 0.3566                          & 0.6627         & 0.6438               & \underline{0.6667}  & \textbf{0.8614}
                    \\ \cmidrule(l){2-11}
                    & \multirow{3}{*}{TT} & Recall   & 0.3774                  & 0.1984                                               & 0.1751      & \underline{0.5331}                    & 0.4747                         & 0.0195     & 0.3891              & \textbf{1.0000}                  \\
                    &                     & Precision & 0.4554                  & 0.3129                                                 & 0.5056                    & \underline{0.5170}                         & 0.3219     & 0.0191   & 0.2801            & \textbf{0.9416}           \\ 
                    &                     & F1 Score & 0.2764                  & 0.3422                                                   & 0.2601                    & \underline{0.5249}                          & 0.3836     & 0.0193   & 0.3257           & \textbf{0.9699}
                    \\ \cmidrule(l){2-11}
                    & \multirow{3}{*}{NT} & Recall   & 0.0860                  & 0.2558                                               & 0.0000   & 0.4256                    & 0.9539                          & \underline{0.9937}     & 0.9686             & \textbf{1.0000}                  \\
                    &                     & Precision & 0.1035                  & 0.1799                                                  & 0.0000                   & 0.1796                          & 0.4892     & \underline{0.4984}   & 0.4920            & \textbf{0.8738}           \\ 
                    &                     & F1 Score & 0.0939                  & 0.2113                                                    & 0.0000                    & 0.2526                         & 0.6468     & \underline{0.6639}  & 0.6525             & \textbf{0.9326}
                    \\ \cmidrule(l){1-11}
\multirow{10}{*}{SP} & \multirow{3}{*}{ST} & Recall   & 0.1863                  & 0.6072                                               & 0.4437    & 0.4410                & \underline{0.9879}                          & 0.9383      & 0.9866              & \textbf{0.9904}                  \\
                    &                     & Precision & 0.3677                  & 0.3826                                                & 0.3177                    & 0.3978                          & \underline{0.4970}     & 0.4961  & 0.4966              & \textbf{0.8573}           \\ 
                    &                     & F1 Score  & 0.2473                  & 0.4694                                                  & 0.3702                    & 0.4183                          & \underline{0.6613}     & 0.6490  & 0.6607             & \textbf{0.9191}
                    \\ \cmidrule(l){2-11}
                    & \multirow{3}{*}{TT} & Recall   & 0.2273                  & 0.3760                                               & 0.4256                    & 0.5744                          & \underline{1.0000}     & 0.2934  & 0.0661              & \textbf{1.0000}                  \\
                    &                     & Precision & 0.4741                  & 0.3792                                                 & \underline{0.5255}                    & 0.4112                          & 0.5000     & 0.2268  & 0.0620             & \textbf{0.9277}           \\ 
                    &                     & F1 Score & 0.3073                  & 0.3776                                                  & 0.4703                    & 0.4793                          & \underline{0.6667}     & 0.2559  & 0.0640             & \textbf{0.9625}
                    \\ \cmidrule(l){2-11}
                    & \multirow{3}{*}{NT} & Recall   & 0.0640                  & 0.1560                                               & 0.0000                    & 0.5800                          & 0.5240     & \underline{0.9880}  & 0.9400            & \textbf{1.0000}                  \\
                    &                     & Precision & 0.0816                  & 0.1204                                                  & 0.0000                    & 0.2562                          & 0.4533     & \underline{0.4970}  & 0.4942             & \textbf{0.6583}           \\ 
                    &                     & F1 Score & 0.0717                  & 0.1359                                                 & 0.0000                    & 0.3554                          & 0.4861     & \underline{0.6613}    & 0.6478           & \textbf{0.7940}
                    \\ \cmidrule(l){1-11}
\multirow{10}{*}{US} & \multirow{3}{*}{ST} & Recall   & 0.1859                  & 0.5283                                               & 0.1249                    & 0.7672                          & 0.9875     & 0.7797    & \underline{0.9875}          & \textbf{1.0000}                  \\
                    &                     & Precision & 0.4457                  & 0.4380                                                & 0.4134                    & \underline{0.5089}                          & 0.4968     & 0.4881  & 0.4968             & \textbf{0.8697}           \\ 
                    &                     & F1 Score & 0.2624                  & 0.4789                                                 & 0.1919                    & 0.6119                          & 0.6611     & 0.6004  & \underline{0.6611}             & \textbf{0.9303}
                    \\ \cmidrule(l){2-11}
                    & \multirow{3}{*}{TT} & Recall   & 0.1599                  & 0.2876                                               & 0.5060                    & \underline{0.5105}                          & 0.0000    & 0.0242  & 0.0178           & \textbf{1.0000}                  \\
                    &                     & Precision & 0.3327                  & 0.2576                                                & \underline{0.5595}                    & 0.4221                          & 0.0000     & 0.0237  &  0.0175           & \textbf{0.7809}           \\ 
                    &                     & F1 Score & 0.2160                  & 0.2718                                                 & \underline{0.5314}                    & 0.4621                          & 0.0000     & 0.0239   &  0.0176          & \textbf{0.8770}
                    \\ \cmidrule(l){2-11}
                    & \multirow{3}{*}{NT} & Recall   & 0.1520                  & 0.1850                                               & 0.4830                    & \underline{0.7990}                          & 0.9440     & 0.9970    & \underline{0.9970}         & \textbf{1.0000}                  \\
                    &                     & Precision & 0.2021                  & 0.1446                                               & 0.1868                    & 0.2706                          & 0.4856     & 0.4992   &  0.\underline{4992}          & \textbf{0.7249}           \\ 
                    &                     & F1 Score & 0.1735                  & 0.1624                                               & 0.2694                    & 0.4042                          & 0.6413     & 0.6653   &  \underline{0.6653}          & \textbf{0.8405}
                    \\ \cmidrule(l){1-11}
\end{tabular}
\end{table*}

\begin{table*}[h]
\caption{Performance comparison in three real-world datasets for behavior prediction.}
\label{tab:combp}
\setlength{\tabcolsep}{0.6em}
% \resizebox{\textwidth}{60mm}{
\begin{tabular}{@{}cccccccccccc@{}}
\toprule
Dataset             & Type                & Metric   & \multicolumn{1}{c}{HMM} & \multicolumn{1}{c}{FPMC} & \multicolumn{1}{c}{LSTM} & \multicolumn{1}{c}{CARNN} & \multicolumn{1}{c}{Caser}  & \multicolumn{1}{c}{SIAR} & \multicolumn{1}{c}{SASRec} & \multicolumn{1}{c}{\schemename} \\ \midrule
\multirow{7}{*}{FR} & \multirow{2}{*}{ST} & NDCG@10   & \underline{0.4349}                  & 0.3073                              & 0.4012                   & 0.3563                 & 
                    0.4060                    & 0.3691                          & 0.2393                    & \textbf{0.5636}                  \\
                    &                     & HR@10 & \underline{0.6306}                  & 0.4234                              & 0.5315                  & 0.5225                    & 0.5135                    & 0.4955                          & 0.4483                     & \textbf{0.9540}                
                    \\ \cmidrule(l){2-11}
                    & \multirow{2}{*}{TT} & NDCG@10   & 0.3947                  & 0.0834                              & 0.3448                  & 0.2304                 & 
                    \underline{0.4087}                    & 0.2685                         & 0.3130                  & \textbf{0.5174}                  \\
                    &                     & HR@10  & 0.5953                  & 0.1634                             & 0.4319                  & 0.4786                   & \underline{0.5953}                    & 0.5798                          & 0.5205                    & \textbf{0.7834}           
                    \\ \cmidrule(l){2-11}
                    & \multirow{2}{*}{NT} & NDCG@10   & 0.2802                  & 0.2562                              & 0.3598                   & 0.3589                 & 
                    \underline{0.4558}                    & 0.3505                          & 0.1846                  & \textbf{0.6571}                  \\
                    &                     & HR@10  & 0.4612                  & 0.3941                              & 0.5241                  & 0.5786                      & \underline{0.6352}                    & 0.5514                          & 0.2697                   & \textbf{0.9213}          
                    \\ \cmidrule(l){1-11}
\multirow{7}{*}{SP} & \multirow{2}{*}{ST} & NDCG@10   & 0.3578                  & 0.2684                              & 0.3587                   & 0.3388                 & 
                    \underline{0.3696}                    & 0.3243                          & 0.1183                    & \textbf{0.7332}                  \\
                    &                     & HR@10  & \underline{0.4759}                  & 0.3928                              & 0.4303                   & 0.4276                   & 0.4290                    & 0.4115                          & 0.1926                    & \textbf{0.9326}          
                    \\ \cmidrule(l){2-11}
                    & \multirow{2}{*}{TT} & NDCG@10   & 0.5387                  & 0.4158                              & \underline{0.5879}                   & 0.4943                 & 
                    0.4754                    & 0.5040                          & 0.2723                  & \textbf{0.6178}                  \\
                    &                     & HR@10  & \underline{0.7231}                  & 0.5041                              & 0.7066                  & 0.6074                   & 0.5702                    & 0.6281                          & 0.6528                   & \textbf{0.8956}         
                    \\ \cmidrule(l){2-11}
                    & \multirow{2}{*}{NT} & NDCG@10   & 0.3554                  & 0.4188                              & 0.5528                   & 0.4930                 & 
                    \textbf{0.5692}                    & 0.4385                          & 0.2752                   & \underline{0.5614}                  \\
                    &                     & HR@10   & 0.4960                  & 0.6580                              & 0.7840                  & 0.7800                    & \underline{0.7880}                    & 0.6660                          & 0.4650                  & \textbf{0.8153}           
                    \\ \cmidrule(l){1-11}
\multirow{7}{*}{US} & \multirow{2}{*}{ST} & NDCG@10   & 0.2864                  & 0.2366                              & 0.2685                   & 0.2427                 & 
                    0.3215                    & 0.1832                          & \underline{0.4696}                   & \textbf{0.7073}                  \\
                    &                     & HR@10    & 0.4435                  & 0.3748                              & 0.4313                  & 0.4287                           & 0.4489                    & 0.4287                          & \underline{0.6362}                   & \textbf{0.9134}           
                    \\ \cmidrule(l){2-11}
                    & \multirow{2}{*}{TT} & NDCG@10   & 0.3226                  & 0.3711                              & 0.4214                   & 0.3772                 & 
                    0.4115                    & 0.3383                          & \underline{0.5379}                  & \textbf{0.7239}                  \\
                    &                     & HR@10    & 0.5506                  & 0.5318                              & 0.6029                  & 0.5544                           & 0.5961                    & 0.5118                          & \underline{0.6626}                    & \textbf{0.9503}          
                    \\ \cmidrule(l){2-11} 
                    & \multirow{2}{*}{NT} & NDCG@10   & 0.2281                  & 0.2250                              & 0.2742                   & 0.2873                 & 
                    0.2905                    & 0.2595                          & \textbf{0.6118}                   & \underline{0.5616}                  \\
                    &                     & HR@10    & 0.3860                  & 0.3340                              & 0.4710                  & 0.4700                           & 0.4940                    & 0.4270                          & \textbf{0.8226}                   & \underline{0.7974}       
                    \\ \cmidrule(l){1-11}
\end{tabular}
\end{table*}

\subsection{Experimental Setup}
%\subsubsection{Datasets}
%\textbf{Datasets}

% We use three real-world datasets consisting of only normal samples, two (FR/SP) from public datasets and one anonymous dataset (AN) collected by ourselves. The datasets description is shown in Table 1. All datasets are split into training, validation and testing sets with a ratio of 7:1:2. 
% We use three real-world datasets consisting of only normal samples (FR/SP/US) from the public datasets. We divide each of these three datasets into 6 datasets: winter, spring, daytime, night, single and multiple according to different seasons, different activity times and different activity frequencies. Among them, winter, daytime, and single are used as original datasets, corresponding to the three context changes of ST, TT, and NT. Spring, night, and multiple are used as tests of the target context, providing complete real-world data support for the verification of experimental results.
We utilize three real-world datasets containing only normal samples (FR, SP, and US) sourced from publicly available repositories. Each dataset is further partitioned into six subsets based on season (winter, spring), data collection period (daytime only, nighttime only), and household occupancy status (single-user vs. multi-user). Among these, the winter, daytime, and single-user subsets are used as original datasets, while the remaining subsets serve as target datasets. These pairs reflect three types of context shift: seasonal transition (ST), time-schedule transition (TT), and user occupancy transition (NT), respectively. %This setup enables comprehensive evaluation of model generalization under realistic behavior drift scenarios.

%\subsubsection{Prompt} 
%\noindent\textbf{Prompt}

The main prompt contents are shown in Table~\ref{tab:findings}, which include role, background, CoT task decomposition, requirements, scene and device information. The LLM used for synthesis is GPT-4o.

%\subsubsection{Baselines}
%\noindent\textbf{Baselines}

% We compare \schemename's capabilities for anomaly detection task with existing general anomaly detection methods in smart homes:
We use data synthesized by \schemename \ to retrain a Transformer Autoencoder for the anomaly detection task to demonstrate the performance of our system. The baselines include several state-of-the-art anomaly detection methods in smart homes:

\textbf{Local Outiler Factor (LOF)} \cite{cheng2019outlier} calculates the density ratio between each sample and its neighbors to detect anomaly.

\textbf{Isolation Forest (IF)} \cite{liu2008isolation} builds binary trees, and instances with short average path lengths are detected as anomaly.
% \item \textbf{6thSense} utilizes Naive Bayes to detect malicious behavior associated with sensors in smart homes.

 \textbf{Aegis} \cite{Siker19Aegis} utilizes a Markov Chain-based machine 
 learning technique to detect malicious behavior in smart homes.

\textbf{OCSVM} \cite{amraoui2021ml} build a One-Class Support Vector Machine model to prevent malicious control of smart home systems.

\textbf{Autoencoder} \cite{chen2018autoencoder} learns to reconstruct normal data and then uses the reconstruction error to determine abnormal.

\textbf{ARGUS} \cite{Rieger23ARGUS} designed an Autoencoder based on Gated Recurrent Units (GRU) to detect IoT infiltration attacks.

\textbf{TransformerAutoencoder (TransAE)} \cite{vaswani2017attention} uses self-attention in the encoder and decoder to achieve anomaly detection.
%\end{itemize}

% We compare \schemename's capabilities for behavior prediction task with existing studies about UDI prediction in smart home and sequential user behavior prediction algorithms:
For behavior prediction task, we use data synthesized by \schemename \ to retrain a SASRec. The baselines include several state-of-the-art behavior prediction algorithms:

\textbf{HMM} \cite{lin2018sequential} builds a transition matrix between different device controls to capture the first order transition probabilities for BP.

\textbf{FPMC} \cite{rendle2010factorizing} combines Markov chain with matrix factorization to capture sequential patterns and user preferences for BP.

\textbf{LSTM} \cite{tax2018human} captures the long-term sequential influence for BP.

\textbf{CARNN} \cite{liu2016context} considers context-dependent features by context-specific transition matrix in RNN for a sequential recommendation.

\textbf{Caser} \cite{tang2018personalized} employs CNN in both time-axis and feature-axis to capture temporal dynamics for BP.
% \item \textbf{DeepMove} captures both long and short-term user behavior patterns by the enhanced version of RNN with a history attention mechanism.

\textbf{SIAR} \cite{rakkappan2019context} applies Stacked Recurrent Neural Networks that model the dynamics of contexts and temporal gaps for BP.
% \item \textbf{SR-GNN} applies gated graph neural network to generate latent vectors of items and then represents each session through an attention network for user behavior prediction.

\textbf{SASRec} \cite{kang2018self} uses time-growing directional transformer encoder to consider sequential patterns of user actions for action prediction.
% \item \textbf{SmartSense} applies query transformer for smart home action recommendation.
% \item \textbf{DeepUDI} applies graph neural network, transformer and attention mechanism to predict user behavior in smart home scenarios.

%\subsubsection{Evaluation Metrics}
%\noindent\textbf{Evaluation Metrics}

% We measure the system's ability to synthesize data by using the performance of the model retrained on the synthetic dataset on two common downstream tasks in smart homes. We use common metrics such as \textit{Recall}, \textit{Precision}, and \textit{F1-Score} to evaluate the performance of Anomaly Detection (AD) task. We use common metrics such as \textit{NDCG@10} and \textit{HR@10} to evaluate the performance of Behavior Prediction (BP) task. And their specific calculation process and explanation are shown in Appendix~\ref{calculation}.
%We evaluate the system’s data synthesis capability by assessing the performance of a model retrained on the synthetic dataset across two representative downstream tasks in smart home. 
For the Anomaly Detection (AD) task, we adopt standard evaluation metrics, including \textit{Recall}, \textit{Precision}, and \textit{F1-Score}. For the Behavior Prediction (BP) task, we use \textit{NDCG@10} and \textit{HR@10} as evaluation metrics. The detailed computation procedures and explanations of these metrics are provided in Appendix~\ref{calculation}.

\begin{table}[H]
\caption{Ablation study of pre-synthesis components on AD and BP tasks (w/o denotes without).}
\label{tab:ablationpre}
\setlength{\tabcolsep}{0.2em}
\begin{tabular}{@{}ccccccc@{}}
\toprule
     Task                    & Metric   & \multicolumn{1}{c}{w/o ALL} & \multicolumn{1}{c}{w/o SSC} & \multicolumn{1}{c}{w/o TSS} & \multicolumn{1}{c}{w/o GSS} & \multicolumn{1}{c}{\schemename} \\ \midrule
\multirow{3}{*}{AD} & Recall   & 0.2121 & 0.1736 & 0.3554 & \underline{1.0000} & \textbf{1.0000} \\
                    & Precision & 0.2750 & 0.4565 & 0.6324 & \underline{0.7790} & \textbf{0.7996}  \\ 
                    & F1 Score & 0.2395 & 0.2515 & 0.4550 & \underline{0.8758} & \textbf{0.8886}  \\
\midrule
\multirow{2}{*}{BP} & NDCG@10   & 0.0774 & \underline{0.4819} & 0.3161 & 0.1523 & \textbf{0.6521}  \\
                    & HR@10 & 0.1458 & \underline{0.9298} & 0.6259 & 0.2297 & \textbf{0.9477}  \\
\midrule
\end{tabular}
\end{table}

\subsection{Performance}

%We use grid search to adjust the parameters of \schemename \ and report
The overall performance of \schemename \ and all baselines are shown in Table~\ref{tab:comad} and Table~\ref{tab:combp}. Bold and underlined values indicate the best and the second best performance. 
\schemename \ outperforms all baselines in most cases, as it effectively synthesizes high-quality data that aligns with the target context, enabling the retrained model to better adapt to environmental shifts. In contrast, models without retraining lack adaptive capabilities and generally exhibit degraded performance. For example, in anomaly detection tasks, they often suffer from low precision due to the presence of novel behavior combinations under the new context that were not observed during initial training. Likewise, behavior prediction models experience reduced success rates as environmental changes cause substantial shifts in user behavior patterns.

\subsection{Ablation Study} \label{ablation}
{\schemename} has four main functional modules, including three pre-synthesis data processing and analysis modules: {\timesplit} ({\TS}), {\compression} ({\cp}), {\graph} ({\gp}), and one post-synthesis moldule, i.e., {\filter} ({\dof}). 
% \begin{table}[h]
% \caption{Ablation study of pre-synthesis components on Anomaly Detection (AD) and Behavior Prediction (BP) (w/o denotes without).}
% \label{tab:ablationpre}
% \setlength{\tabcolsep}{0.2em}
% \begin{tabular}{@{}ccccccc@{}}
% \toprule
%      Task                    & Metric   & \multicolumn{1}{c}{w/o ALL} & \multicolumn{1}{c}{w/o SSC} & \multicolumn{1}{c}{w/o TSS} & \multicolumn{1}{c}{w/o GSS} & \multicolumn{1}{c}{\schemename} \\ \midrule
% \multirow{3}{*}{AD} & Recall   & 0.0110 & 0.1928 & 0.1873 & \textbf{1.0000} & \textbf{1.0000} \\
%                     & Precision & 0.0225 & 0.4930 & 0.4857 & 0.7691 & \textbf{0.7857}  \\ 
%                     & F1 Score & 0.0148 & 0.2772 & 0.2704 & 0.8695 & \textbf{0.88}  \\
% \midrule
% \multirow{2}{*}{BP} & NDCG@10   & 0.1274 & \textbf{0.6254} & 0.4053 & 0.1497 & 0.6026  \\
%                     & HR@10 & 0.1884 & 0.9230 & 0.6396 & 0.2105 & \textbf{0.9395}  \\
% \midrule
% \end{tabular}
% \end{table}

To investigate different pre-synthesis components’ effectiveness in \schemename, we implement 5 variants of \schemename \ for ablation study. 
As shown in Table~\ref{tab:ablationpre}, each pre-synthesis component of \schemename \ has a positive impact on results. The combination of all components brings the best overall results, which is much better than using any subset of the three components. Each module fulfills a role aligned with its designated function. 

% At the same time, it can be seen that {\TS} and {\cp} enhance the system's understanding of data information, while {\gp} strengthens the retention of user characteristics and improves its performance in behavior prediction.
\begin{table}[H]
\caption{Ablation study of the post-synthesis component ({\dof}) on AD (Number means the number of sequences in the sythesized dataset).}
\label{tab:ablationpost}
\setlength{\tabcolsep}{0.5em}
\begin{tabular}{@{}cccc@{}}
\toprule
     Dataset                 & Type   & \multicolumn{1}{c}{Number ($Ori$/$S1$/$S2$)} & \multicolumn{1}{c}{F1 Score ($Ori$/$S1$/$S2$)}  \\ \midrule
\multirow{3}{*}{FR} & ST   & 137/118/125 & 0.8529/0.8515/\textbf{0.8614}  \\
                    & TT & 28/27/28 & 0.9699/0.9699/0.9699   \\ 
                    & NT  & 52/47/47 & 0.9326/0.9326/0.9326  \\
\midrule
\multirow{3}{*}{SP} & ST   & 146/132/140 & 0.8941/0.9031/\textbf{0.9191} \\
                    & TT & 52/46/52 & \textbf{0.9625}/0.7862/\textbf{0.9625} \\
                    & NT & 100/87/89 & 0.0000/0.7861/\textbf{0.7940}  \\
\midrule
\multirow{3}{*}{US} & ST   & 99/86/94 & 0.8623/0.8838/\textbf{0.9303} \\
                    & TT  & 89/77/84 & 0.8279/0.8055/\textbf{0.8770}  \\
                    & NT  & 214/189/204 & 0.8064/0.8337/\textbf{0.8405}  \\
\midrule
\end{tabular}
\end{table}

Table~\ref{tab:ablationpost} presents the ablation results of the {\dof} module, which consists of two sequential stages. $Ori$ denotes the raw synthetic data without any post-processing, $S1$ indicates the result after applying only the first stage of {\dof}, and $S2$ includes both stages. The improvements observed from $Ori$ to $S1$ and further to $S2$ demonstrate that {\dof} significantly improves data quality by effectively removing implausible or anomalous samples. These results highlight the necessity and effectiveness of the {\dof} module.

\subsection{Case Study}
\begin{figure}[ht]
    \centering
    \includegraphics[width = .4\textwidth]{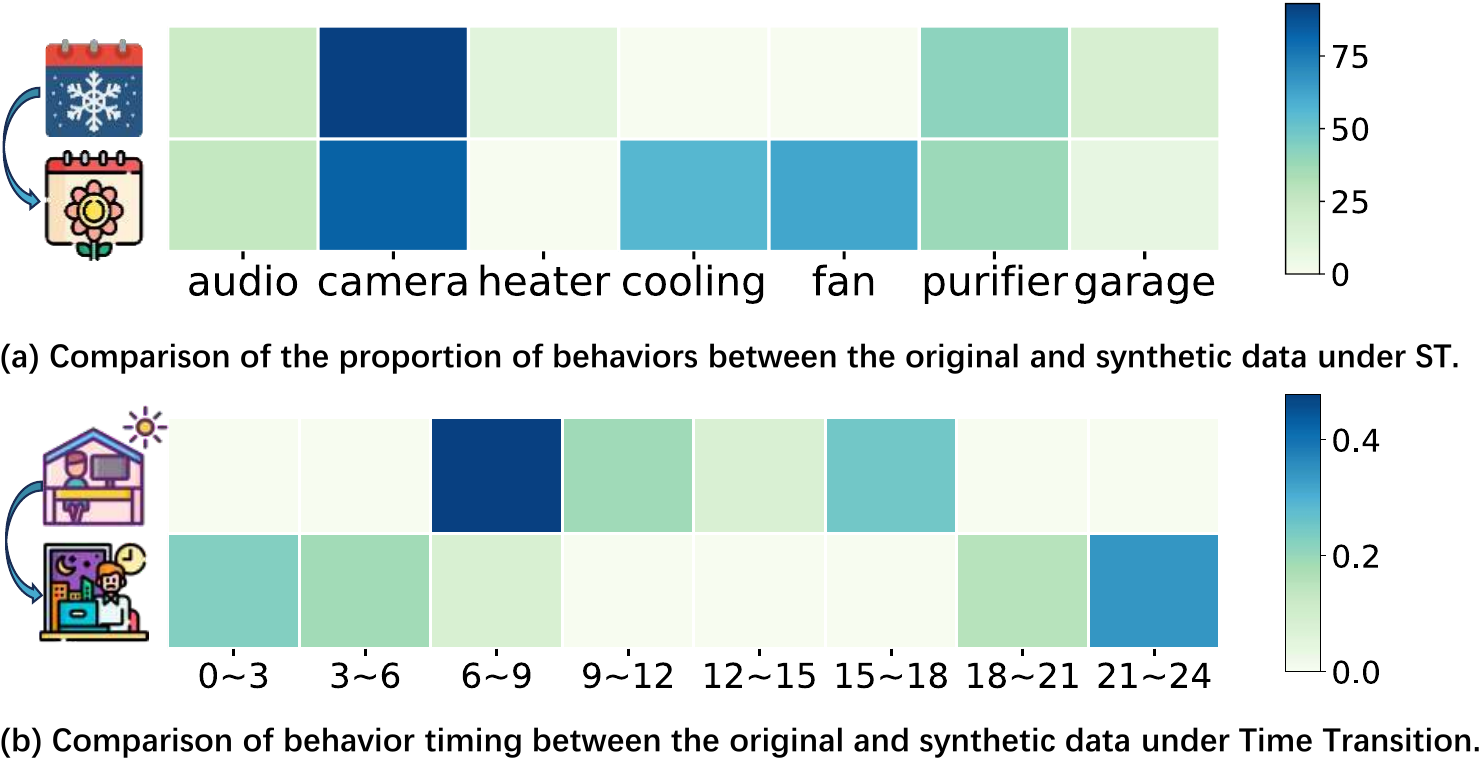}
    \caption{Comparison between the original data and synthetic data on FR.}
    \label{case}
\end{figure}

We select two cases on the FR dataset to visually demonstrate the effect of \schemename . Figure~\ref{case}(a) illustrates the behavior distribution in the original dataset and the synthetic data generated by \schemename\ in response to a ST from winter to summer. As shown, \schemename\ effectively adapts to the new context by increasing the use of air conditioners and fans while removing heater usage. Meanwhile, it largely preserves the original distribution of behaviors unrelated to weather.
Figure~\ref{case}(b) shows \schemename \ effectively captures the shift in the user's daily routine, adapting from daytime activities to nighttime activities.

\section{Conclusion}
In this paper, we introduce \schemename \ for context-aware adaptive synthesis of user behavior. We first devise a {\timesplit} module to reasonably segment overlong sequences while maintaining semantic coherence. Then, we introduce a {\compression} method to perform efficient compression of the segmented subsequences based on semantic space mapping. Meanwhile, we design a {\graph} module to construct a relationship graph to record global user behavior regularities and form a guideline to reduce global information loss caused by compression. Furthermore, we design a {\filter} method to remove noise and anomalies from synthetic data. Comprehensive experiments conducted on three real datasets demonstrate that \schemename \ consistently outperforms baselines while delivering highly interpretable results.
% In this paper, we introduce \schemename \ for unsupervised user behavior anomaly detection. We first devise a Loss-guided Dynamic Mask Strategy (LDMS) to encourage the model to learn less frequent behaviors that are often overlooked during the learning process. Additionally, we introduce Three-level Time-aware Position Embedding (TTPE) to integrate temporal information into positional embedding, allowing for the detection of temporal context anomalies. Furthermore, we propose a Noise-aware Weighted Reconstruction Loss (NWRL) to assign distinct weights for routine behaviors and noise behaviors, thereby mitigating the impact of noise. Comprehensive experiments conducted on three datasets encompassing ten types of anomaly behaviors demonstrate that \schemename \ consistently outperforms state-of-the-art baselines while delivering highly interpretable results.

% \section{Appendices}
% If your work needs an appendix, add it before the
% ``\verb|\end{document}|'' command at the conclusion of your source
% document.

% Start the appendix with the ``\verb|appendix|'' command:
% \begin{verbatim}
%   \appendix
% \end{verbatim}

\balance

\bibliography{sample-base}
\bibliographystyle{ACM-Reference-Format}

\newpage

\appendix

\section{Appendices}

\subsection{Device information of different dataset}
The FR, SP and US data sets contain 33, 34, and 40 devices respectively, as shown in Table~\ref{tab:devInfo_fr}, Table~\ref{tab:devInfo_sp}, and Table~\ref{tab:devInfo_us}.

\begin{table}[h]
\caption{Device information on FR dataset.}
\setlength{\tabcolsep}{.10em}{
\label{tab:devInfo_fr} 
\begin{tabular}{cc|cc|cc}
\hline
No. & Device                  & No. & Device         & No. & Device           \\ \hline
0   & AirConditioner          & 11  & Fan            & 22  & Refrigerator        \\
1   & AirPurifier             & 12  & GarageDoor     & 23  & RemoteController \\
2   & Blind                   & 13  & Light          & 24  & RobotCleaner           \\
3   & Camera                  & 14  & Microwave      & 25  & Siren      \\
4   & ClothingCareMachine     & 15  & MotionSensor   & 26  & SmartLock     \\
5   & Computer                & 16  & NetworkAudio   & 27  & SmartPlug               \\
6   & ContactSensor           & 17  & None           & 28  & Switch      \\
7   & CurbPowerMeter          & 18  & Other          & 29  & Television      \\
8   & Dishwasher              & 19  & Oven           & 30  & Heater  \\
9  & Dryer                    & 20  & PresenceSensor & 31  & Washer      \\
10  & Elevator                & 21  & Projector      & 32  & WaterValve      \\ \hline
\end{tabular}}
\end{table}

\begin{table}[h]
\caption{Device information on SP dataset.}
\setlength{\tabcolsep}{.10em}{
\label{tab:devInfo_sp} 
\begin{tabular}{cc|cc|cc}
\hline
No. & Device                  & No. & Device         & No. & Device           \\ \hline
0   & AirConditioner          & 12  & GarageDoor     & 24  & RobotCleaner         \\
1   & AirPurifier             & 13  & Light          & 25  & SetTop \\
2   & Blind                   & 14  & Microwave      & 26  & Siren            \\
3   & Camera                  & 15  & MotionSensor   & 27  & SmartLock     \\
4   & ClothingCareMachine     & 16  & NetworkAudio   & 28  & SmartPlug    \\
5   & Computer                & 17  & None           & 29  & Switch               \\
6   & ContactSensor           & 18  & Other          & 30  & Television     \\
7   & CurbPowerMeter          & 19  & Oven           & 31  & Heater      \\
8   & Dishwasher              & 20  & PresenceSensor & 32  & Washer  \\
9  & Dryer                    & 21  & Projector      & 33  & WaterValve      \\
10  & Elevator                & 22  & Refrigerator                        \\ 
11  & Fan                     & 23  & RemoteController                    \\ \hline
\end{tabular}}
\end{table}

\begin{table}[h]
\caption{Device information on US dataset.}
\setlength{\tabcolsep}{.05em}{
\label{tab:devInfo_us} 
\begin{tabular}{cc|cc|cc}
\hline
No. & Device                  & No. & Device         & No. & Device           \\ \hline
0   & AirConditioner          & 13  & Irrigation     & 26  & RemoteController         \\
1   & AirPurifier             & 14  & LeakSensor          & 27  & RobotCleaner \\
2   & Blind                   & 15  & Light      & 28  & SecurityPanel            \\
3   & Camera                  & 16  & LightSensor   & 29  & Siren     \\
4   & ClothingCareMachine     & 17  & Microwave   & 30  & SmartLock    \\
5   & Computer                & 18  & MotionSenser         & 31  & SmartPlug               \\
6   & ContactSensor           & 19  & MultiFSensor      & 32  & SmokeDetector     \\
7   & Dishwasher          & 20  & NetworkAudio           & 33  & SoundSensor      \\
8   & Dryer              & 21  & None                & 34  & Switch  \\
9  & Elevator                    & 22  & Other      & 35  & Television      \\
10  & Fan                & 23  & PresenceSensor             & 36 & Heater           \\ 
11  & GarageDoor                     & 24  & Projector   & 37 & Vent   \\
12  & Humidifier         & 25 & Refrigerator      & 38 & Washer   \\
&                & &                              & 39 & WaterValve
\\ \hline
\end{tabular}}
\end{table}

\subsection{Comparison of the effects of different compression methods}
We conduct an experiment to compare the effects of {\cp}. 
% We conduct an experiment to illustrates the drawback of similarity-based method. 
%We set up a basic experiment that is common in anomaly detection tasks: use a training set consisting of normal samples to train an anomaly detection model, and then test the reconstruction error of the model for normal samples in the test set to evaluate the learning effect of the model. We use the entire dataset, the compressed dataset obtained by the similarity method, and the compressed dataset obtained by our \cp method as training sets for the above experiments, and use the same test set for experiments.
We designed a basic experiment that is common in anomaly detection tasks. The goal of this experiment is to train an anomaly detection model using normal samples and evaluate the learning effect of the model. Three different training sets were used in the experiment: Original dataset: The entire dataset was used as the training set; Compressed dataset (similarity method): The compressed dataset obtained by the similarity method was used as the training set; Compressed dataset ({\cp} method): The compressed dataset obtained by the {\cp} method we proposed was used as the training set. Then use one of the three training sets mentioned above to train the anomaly detection model. During the testing phase, the model is evaluated using the same test set, which only contains normal samples. The learning effect of the model is evaluated by calculating the reconstruction loss of the model for normal samples on the test set. A lower reconstruction loss indicates that the model has a better learning effect on normal samples.

% \begin{figure}[ht]
%     \subfigure[Mean construction loss of different compression methods.]{
%     \label{compare1}
%     \centering
%     \includegraphics[width = .31\textwidth]{figure/fr_loss_compare.pdf}
%     }
%     \subfigure[Loss variance of different compression methods.]{
%     \label{compare2}
%     \centering
%     \includegraphics[width = .31\textwidth]{figure/fr_variance_compare.pdf}
%     }
%     \subfigure[Sensitivity of different compression methods to the threshold.]{
%     \label{compare3}
%     \centering
%     \includegraphics[width = .31\textwidth]{figure/fr_number_compare.pdf}
%     }
%     \caption{Performance of different compression methods on the test dataset.}
%     \label{compare_compression}
% \end{figure}

\begin{figure}[ht]
    \subfigure[Mean construction loss of different compression methods.]{
    \label{compare1}
    \centering
    \includegraphics[width = .22\textwidth]{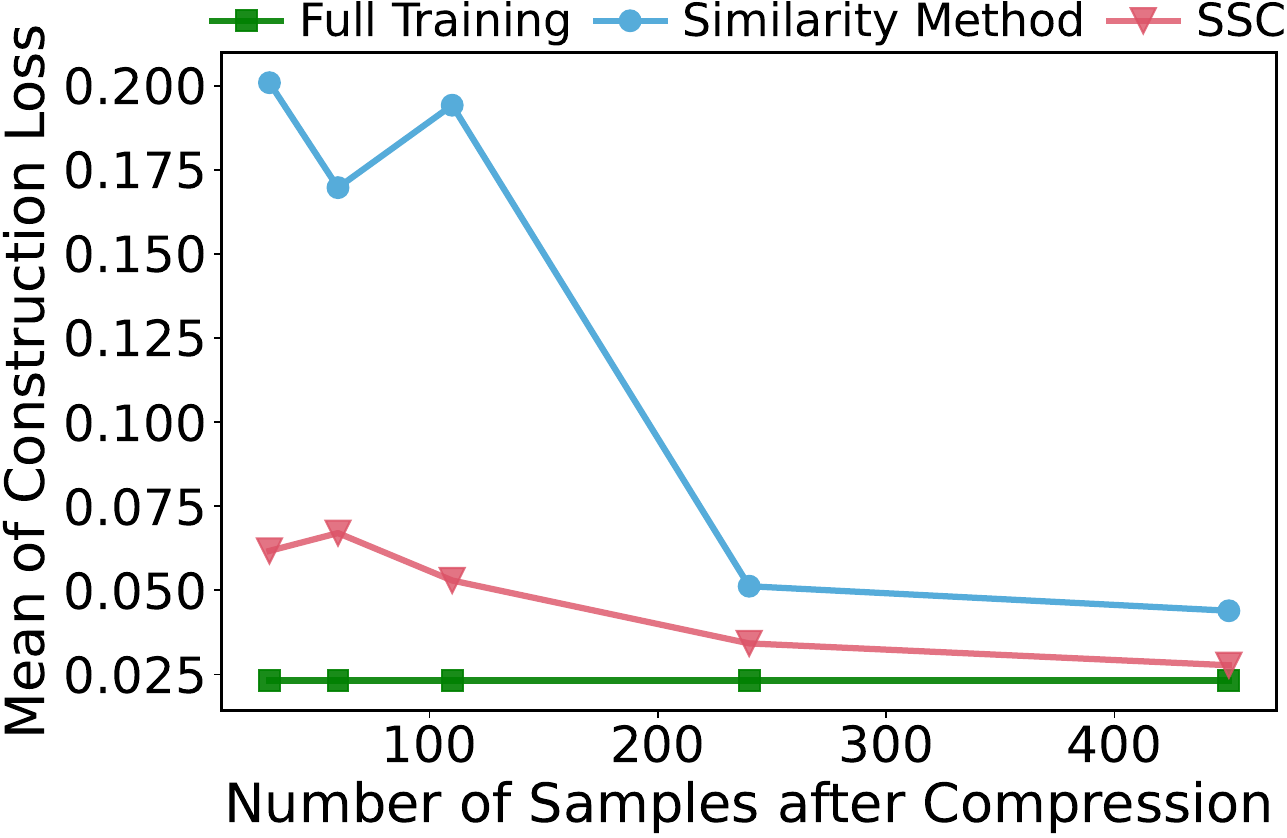}
    }
    \subfigure[Loss variance of different compression methods.]{
    \label{compare2}
    \centering
    \includegraphics[width = .22\textwidth]{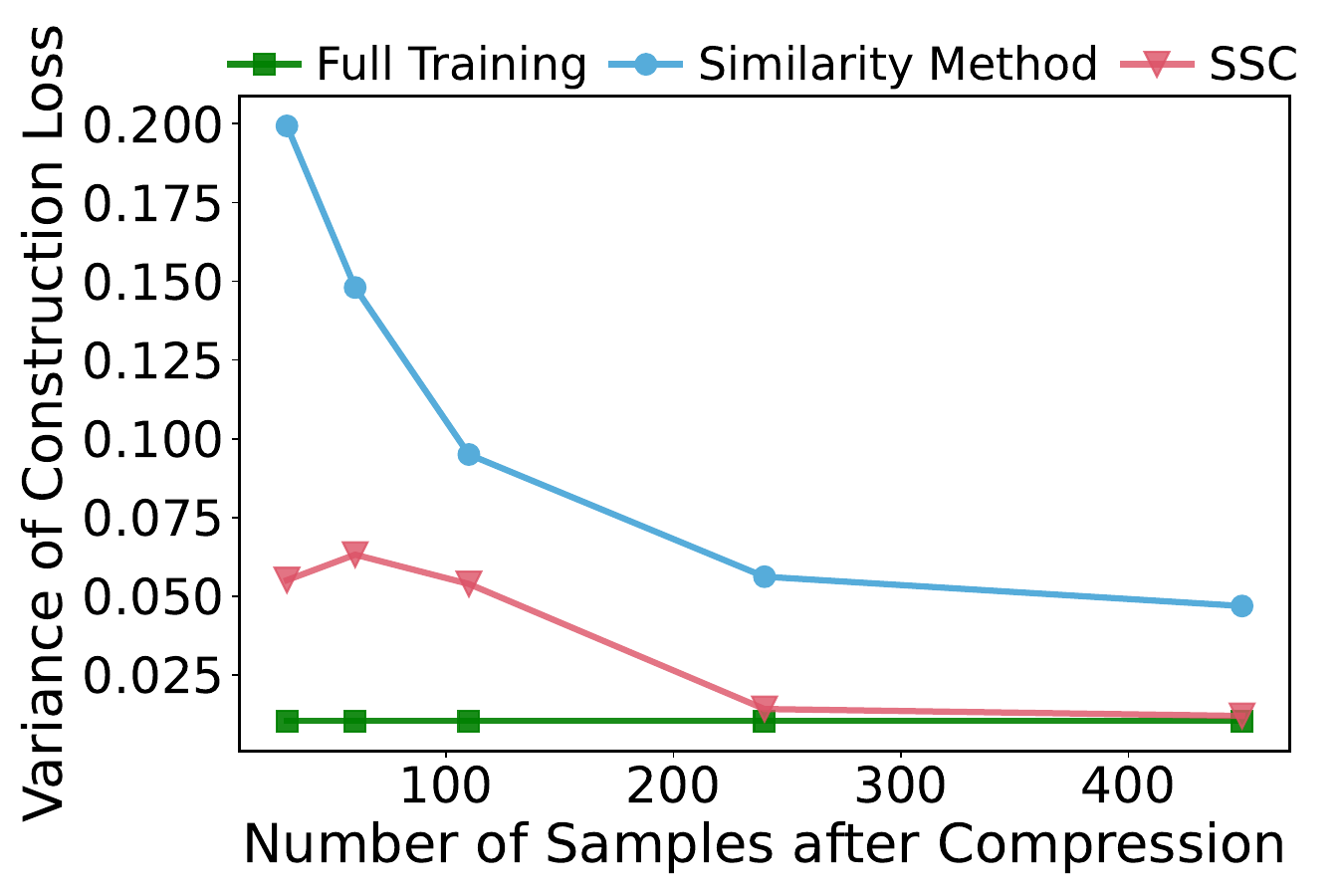}
    }
    \caption{Performance of different compression methods on the test dataset.}
    \label{compare_compression}
\end{figure}

\begin{figure}[ht]
\centering
\includegraphics[width = .4\textwidth]{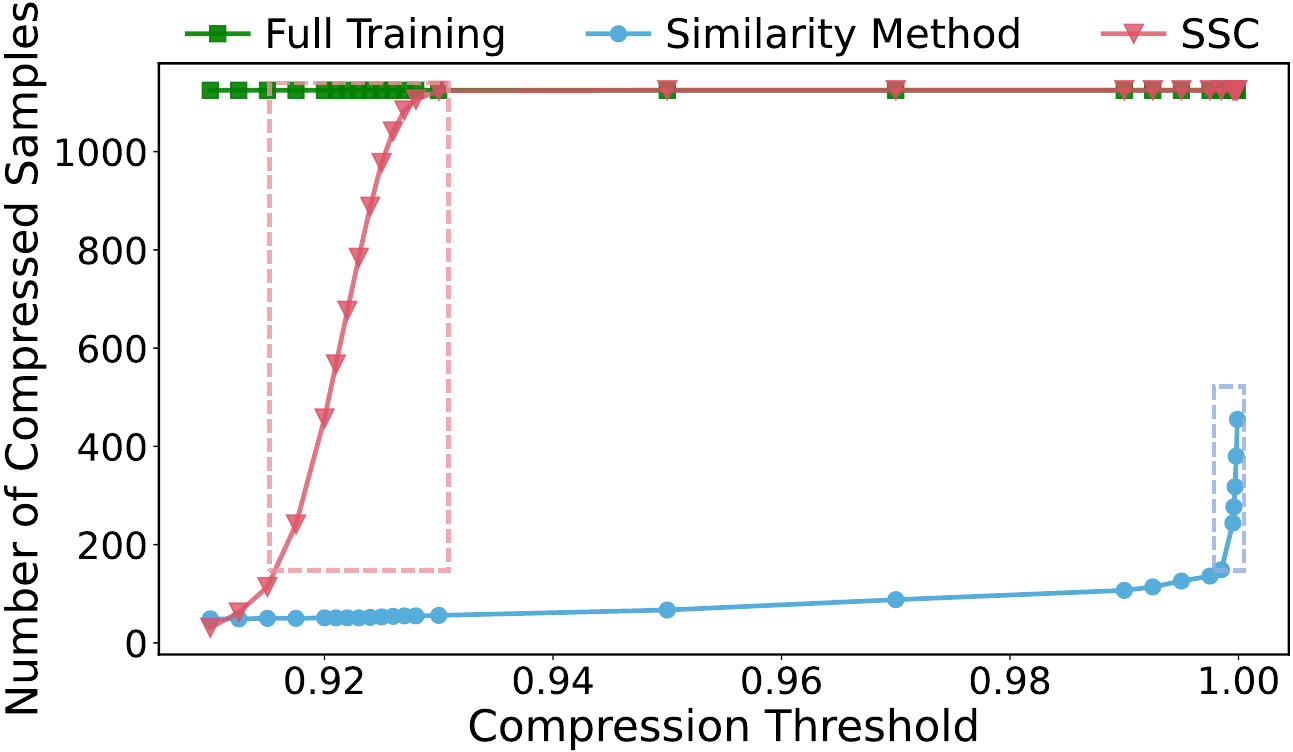}
\caption{Sensitivity of different compression methods to the threshold.}
\label{compare3}
\end{figure}

% As shown in Figure~\ref{compare_compression}, {\cp} compression better preserves the information in the original data set, has smaller fluctuations and lower errors, and can achieve the same performance as the entire original data when the compression rate is 20\%. At the same time, in Figure~\ref{compare3}, the {\cp} method has a larger sensitivity range for the similarity threshold, which can better distinguish and measure each data.
As illustrated in Figure~\ref{compare_compression}, the {\cp} compression method more effectively preserves the information in the original dataset, exhibiting smaller fluctuations and lower errors. It can achieve performance comparable to that of the full original dataset at a 20\% compression rate. Additionally, as shown in Figure~\ref{compare3}, the {\cp} method demonstrates a broader sensitivity range for the similarity threshold, enabling more accurate differentiation and measurement of each data.

\subsection{The compression effect of SSC.} 
We counted and calculated the compression rate and total token savings of the SSC module in two smart home task experiments. The results are shown in the following table: 
\begin{table}[h]
\caption{Sample reduction rate and number of tokens saved by the SSC module.}
\label{tab:ssc_compression}
\setlength{\tabcolsep}{0.8em}
\begin{tabular}{@{}cccc@{}}
\toprule
     Dataset                 & Type   & \multicolumn{1}{c}{Reduction Rate} & \multicolumn{1}{c}{Saved Tokens}  \\ \midrule
\multirow{3}{*}{FR} & ST   & -58.53\% & 136K  \\
                    & TT & -74.44\% & 35K   \\ 
                    & NT  & -84.23\% & 216K  \\
\midrule
\multirow{3}{*}{SP} & ST   & -89.45\% & 786K \\
                    & TT & -71.33\% & 165K \\
                    & NT & -69.29\% & 129K  \\
\midrule
\multirow{3}{*}{US} & ST   & -95.43\% & 1793K \\
                    & TT  & -0\% & 0K  \\
                    & NT  & -29.73\% & 87K  \\
\midrule
\end{tabular}
\end{table}

Table~\ref{tab:ssc_compression} demonstrates that SSC effectively compresses the original data of each dataset to varying degrees while maintaining the quality of data synthesis, achieving an average reduction rate of 63.60\%. The token savings include both the reduction in sample tokens and the tokens associated with additional prompts generated from multiple API calls, resulting in an average total savings of approximately 372K tokens.

\subsection{Computation procedures and explanations of evaluation metrics.} \label{calculation}
We use common metrics such as \textit{Recall}, \textit{Precision}, and \textit{F1-Score} to evaluate the performance of Anomaly Detection (AD) task:
\begin{equation}
    \text { Recall }=\frac{\text { TP }}{\text { TP + FN }}, \ \text { Precision }=\frac{\text { TP }}{\text {  TP + FP }}
\end{equation}
Where $\text {TP}$ is the number that the model correctly identifies anomalies. $\text {FP}$ is the number that the model incorrectly identifies normal instances as anomalies. $\text {TN}$ is the number that the model correctly identifies normal instances. $\text {FN}$ is the number that the model incorrectly identifies anomalies as normal.
\begin{equation}
    \text { F1-Score }=2 \times \frac{\text { Precision } \times \text { Recall }}{\text { Precision }+ \text { Recall }}
\end{equation}
where \text {F1-Score} is the harmonic mean of Precision and Recall, providing a single metric that balances both.

We use common metrics such as \textit{NDCG@10} and \textit{HR@10} to evaluate the performance of Behavior Prediction (BP) task:
\begin{equation}
    \mathrm{DCG} @ 10=\sum_{i=1}^{10} \frac{2^{\mathrm{rel}_i}-1}{\log _2(i+1)}, \ \text { NDCG@10 }=\frac{\text { DCG@10 }}{\text { IDCG@10 }}
\end{equation}
where \text {NDCG@10} evaluates the quality of the top-10 recommendations by considering the position of relevant items in the ranked list. $\mathrm{rel}_i$ is the relevance of the item at position $i$ (e.g., 1 if relevant, 0 otherwise). And \text {IDCG} is the \text {DCG} of the ideal ranking (i.e., the best possible ranking of relevant items). 
\begin{equation}
    \text { HR@10 }=\frac{\text { Numbers of at least one relevant item in top-10 }}{\text { Total recommendations numbers}}
\end{equation}
where \text {HR@10} measures whether at least one relevant item is present in the top-10 recommendations.

\subsection{Parameter Study (RQ3)}
\label{para}
% \subsubsection{The impact of different compression threshold.}
% We select three different compression thresholds $\alpha$: 0.918, 0.919, and 0.920 for generation and test them on two downstream tasks.

% \begin{figure}[ht]
%     \subfigure[The impact of different $\alpha$ on AD.]{
%     \label{thresholdEPAD}
%     \centering
%     \includegraphics[width = .22\textwidth]{figure/thresholdEPAD.pdf}
%     }
%     \subfigure[The impact of different $\alpha$ on BP.]{
%     \label{thresholdEPBP}
%     \centering
%     \includegraphics[width = .22\textwidth]{figure/thresholdEPBP.pdf}
%     }
%     \caption{The Impact of Different Thresholds on Two Tasks.}
%     \label{thresholdEP}
% \end{figure}
% As shown in Figure~\ref{thresholdEP}, as compression threshold is relaxed, the task performance improves gradually across most datasets, as more seed data becomes available. However, when the compression threshold is too stringent, a performance decline may occur.

\subsubsection{Impacts of different split time intervals.}
To control sequence segmentation, we set $\Delta t_{\text{max}}$ (maximum time gap between actions) to 6, 9, 12, and 15 hours, and $T_{\text{max}}$ (maximum total sequence duration) to 12, 24, 36, and 48 hours.
\begin{figure}[ht]
    \subfigure[The impact of different $\Delta t_{max}$.]{
    \label{splitEPAD}
    \centering
    \includegraphics[width = .22\textwidth]{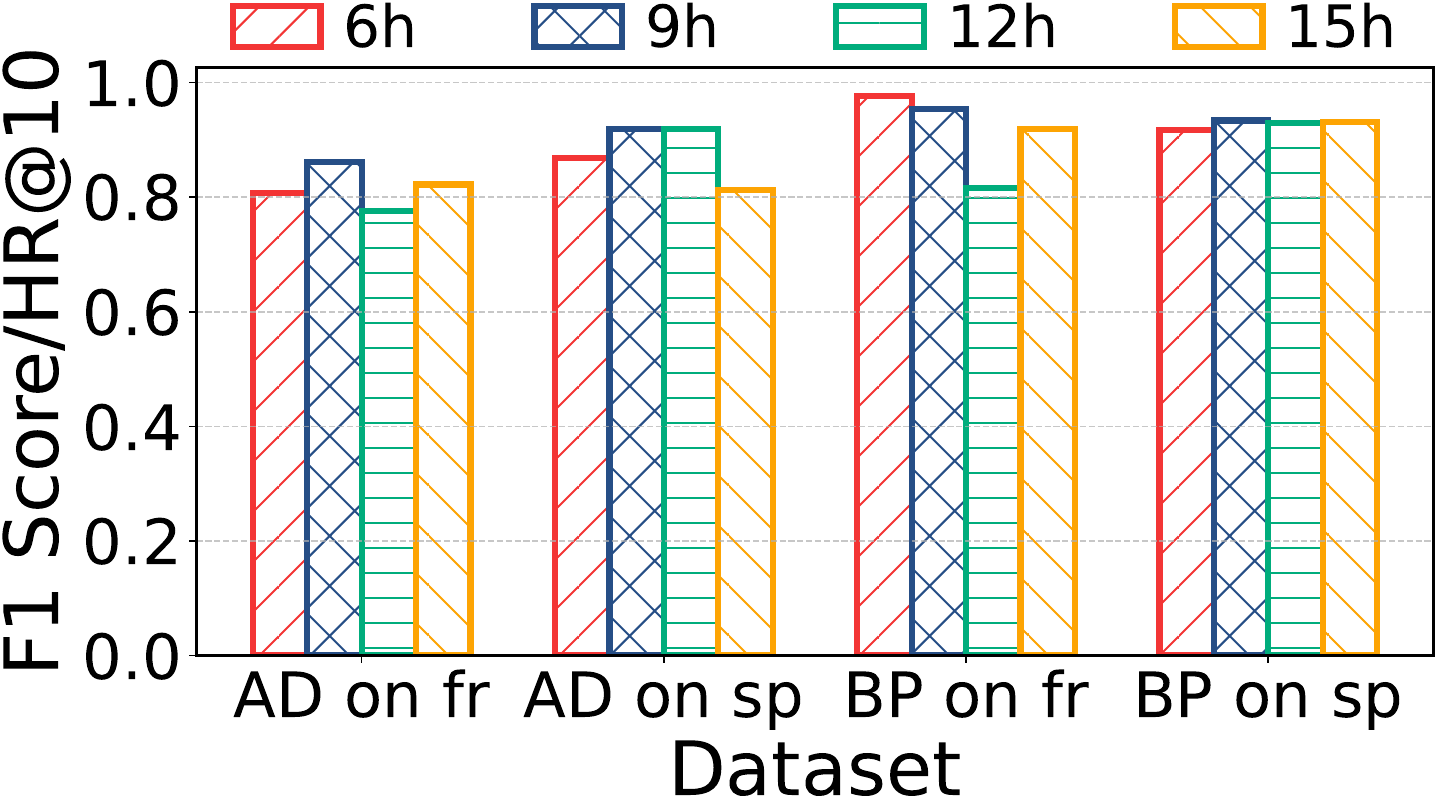}
    }
    \subfigure[The impact of different $T_{max}$.]{
    \label{splitEPBP}
    \centering
    \includegraphics[width = .22\textwidth]{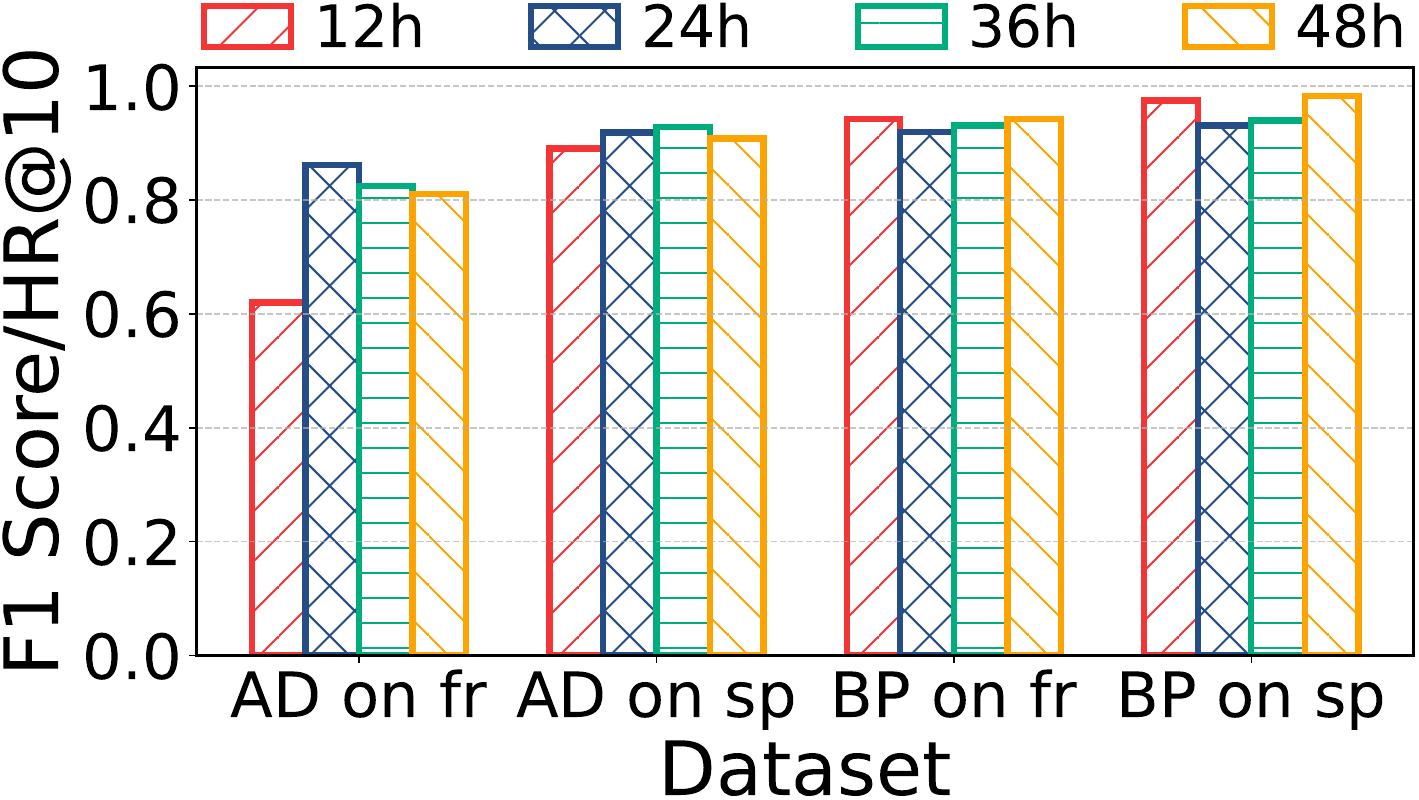}
    }
    \caption{Impacts of different sequence splitting configurations on two tasks.}
    \label{splitEP}
\end{figure}

% As shown in Figure~\ref{splitEP}, a moderate time interval yields better synthesis performance, whereas overly short or long intervals may degrade quality. This is because overly short intervals may yield fragmented sequences lacking sufficient context, while overly long intervals risk merging unrelated behaviors, introducing noise. A moderate interval balances context richness and coherence, leading to higher-quality synthesis. 
As shown in Figure~\ref{splitEP}, a moderate time interval yields better synthesis performance, whereas overly short or long intervals may degrade quality. This is because overly short intervals may yield fragmented sequences lacking sufficient context, while overly long intervals risk merging excessively complex sequence. A moderate interval balances context richness and coherence, leading to higher-quality synthesis. 
%The overall stable results across settings further demonstrate the robustness of {\schemename}.

\subsubsection{Impacts of different LLMs.}
We select three different LLMs: Llama-70B, Qwen2.5-72B and GPT-4o, to study the impact of different LLMs on the quality of synthesized data, and evaluate their performance across two downstream tasks.
\begin{figure}[ht]
    \subfigure[Different Performance on AD.]{
    \label{diffllmEPAD}
    \centering
    \includegraphics[width = .22\textwidth]{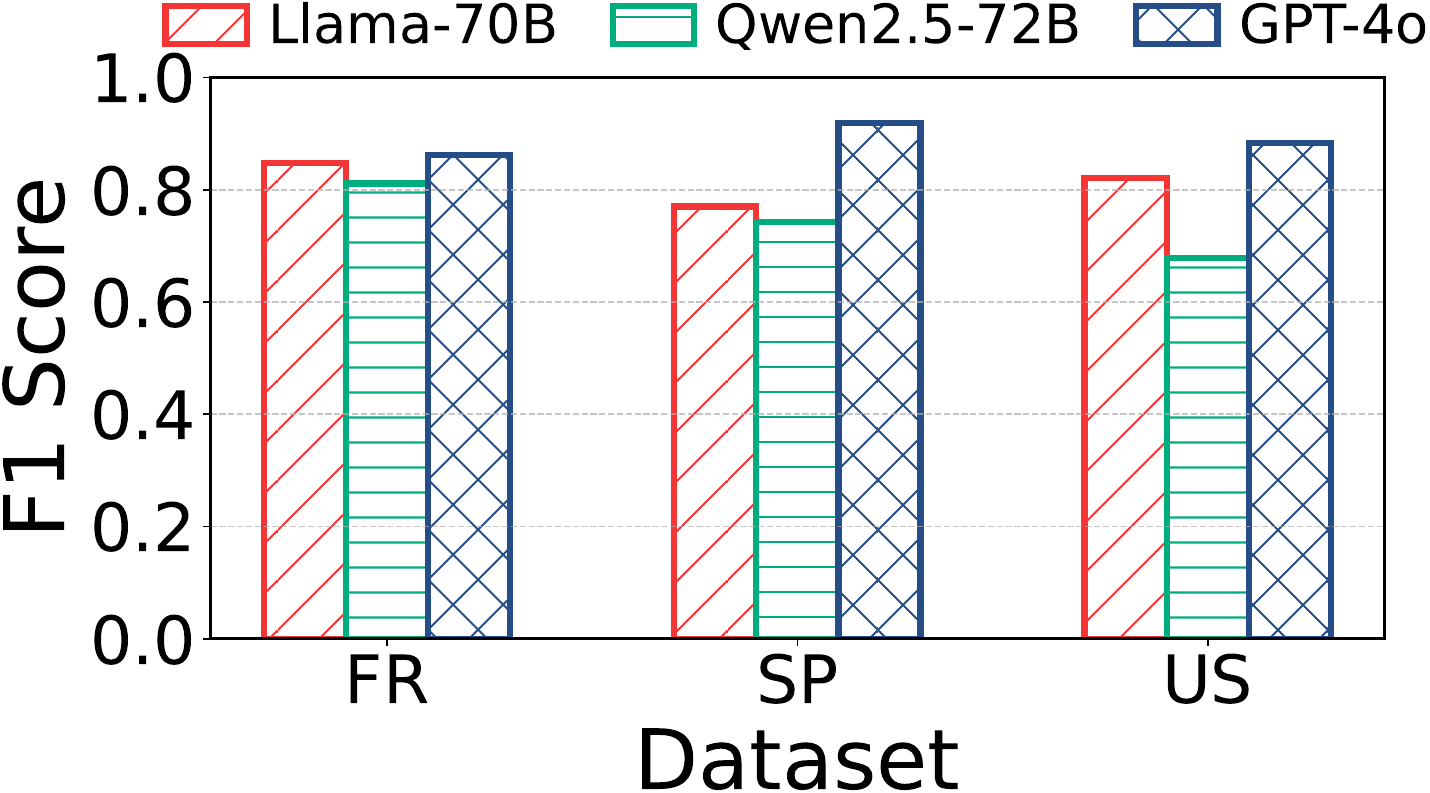}
    }
    \subfigure[Different Performance on BP.]{
    \label{diffllmEPBP}
    \centering
    \includegraphics[width = .22\textwidth]{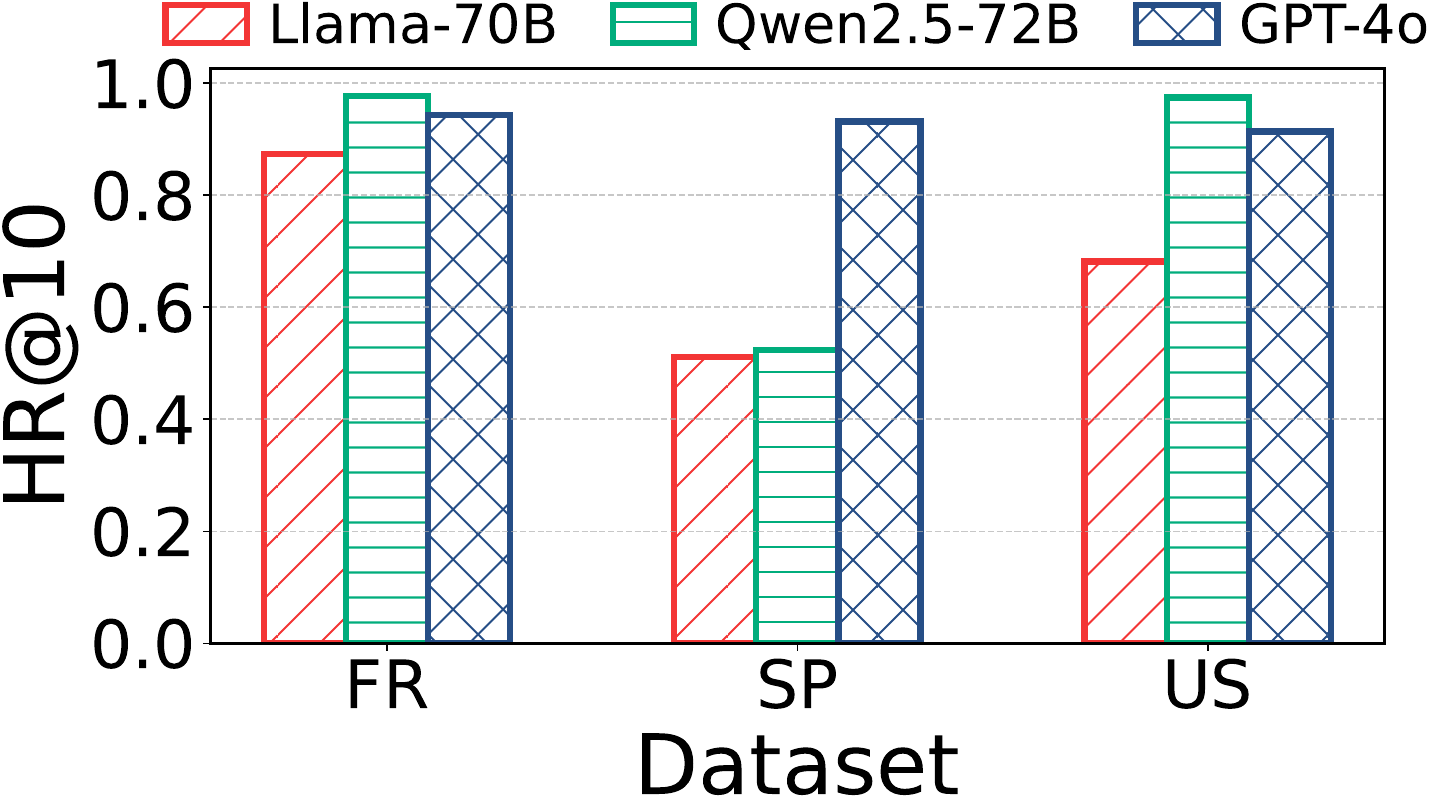}
    }
    \caption{Impacts of different LLMs on two tasks.}
    \label{diffllmEP}
\end{figure}
As shown in Figure~\ref{diffllmEP}, GPT-4o demonstrates superior data synthesis capability, generating behavioral sequences that more closely reflect real-world contextual changes.

\end{document}